\documentclass{article}
\usepackage[preprint]{neurips_2026}
\usepackage[utf8]{inputenc}
\usepackage[T1]{fontenc}
\usepackage{microtype}
\usepackage{url}
\usepackage{amsmath,amssymb,graphicx,xcolor,booktabs,placeins,wrapfig,subcaption}
\usepackage{algorithm}
\usepackage{algpseudocode}
\graphicspath{{./pics/}}
\usepackage{hyperref}
\usepackage{enumitem}

\title{Looking Ahead and Back: Anticipatory Generative Modeling for Road-Network Trajectories}
\author{%
  Zhenhao Huang\textsuperscript{1} \quad
  Li Sun\textsuperscript{2} \quad
  Philip S. Yu\textsuperscript{3} \\
  \textsuperscript{1}North China Electric Power University \\
  \textsuperscript{2}Beijing University of Posts and Telecommunications \\
  \textsuperscript{3}University of Illinois Chicago
}
\begin{document}
\maketitle
\begin{abstract}
Trajectory generation underpins urban applications such as traffic management and urban planning. Within road networks, vehicle movement along road segments is largely deterministic, whereas transitions at intersections are critical—multiple outgoing paths may appear equally plausible, yet they can lead to substantially different downstream routes. Existing generative methods such as diffusion models simulate location sequences and fail short of next transition selection, even producing trajectories that drift off the road network. To fill this gap, we advocate a fundamental shift: moving beyond location simulation toward behavior modeling on the road network. In this paper, we revisit trajectory generation through the lens of generative route planning, and introduce a new graph-constrained generative paradigm, Anticipatory Generative Modeling, which resolves next transition selection by incorporating the \emph{Anticipation} of how the future route may unfold toward the destination. Concretely, we propose a novel generative model named \textsc{Foresight} that formulates a probabilistic model for “\emph{anticipate--select--update--re-anticipate}” loop that enables longer-term foresight while continuing sequential generation along the historical trajectory.
Extensive experiments demonstrate that \textsc{Foresight} wins state-of-the-art baselines. 
\end{abstract}
\section{Introduction}
\label{sec:introduction}
\begin{wrapfigure}{r}{0.55\linewidth}
    \centering
        \vspace{-0.2in}
    \begin{subfigure}{0.49\linewidth}
        \centering
        \includegraphics[width=\linewidth]{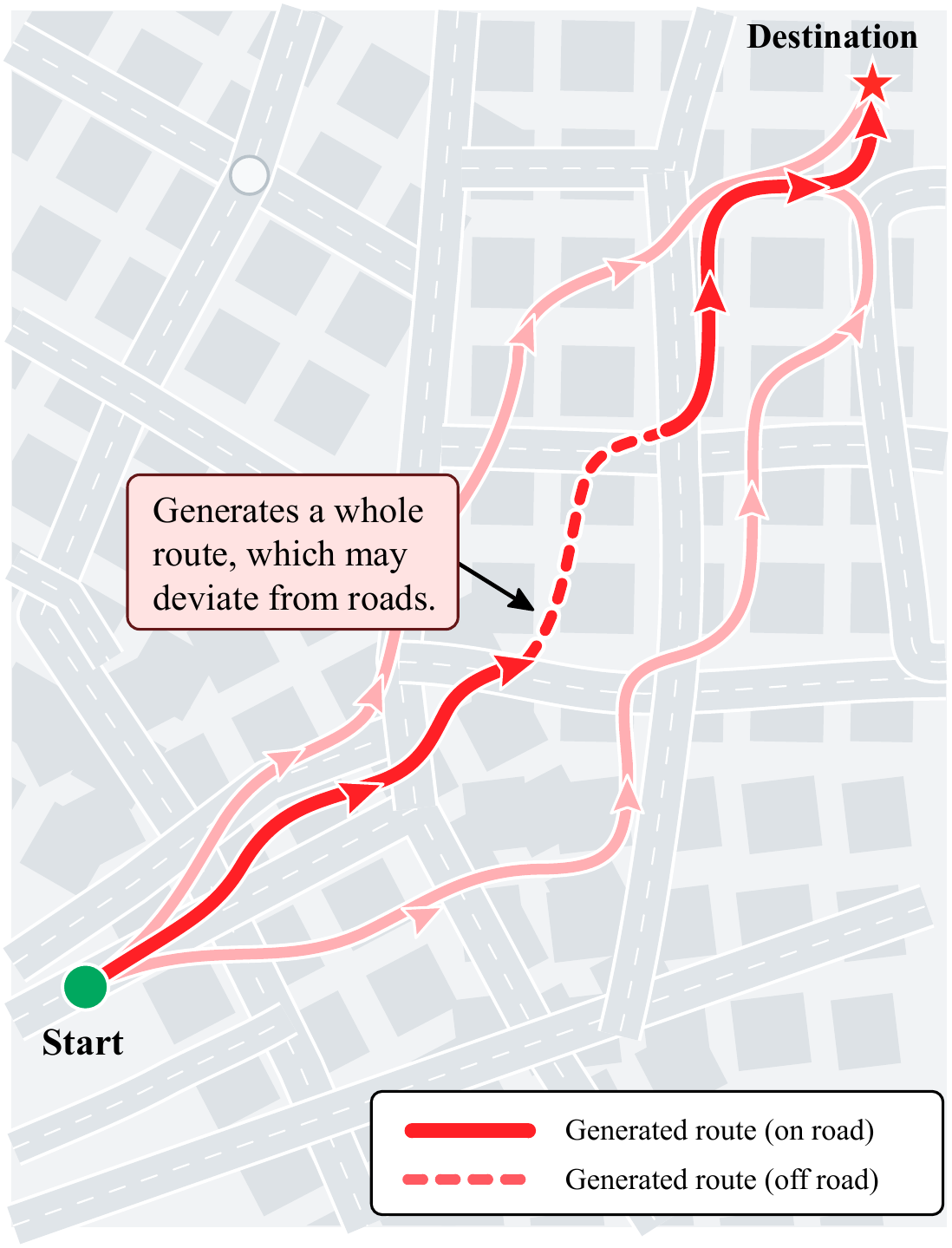}
        \caption{Diffusion model: global route generation.}
        \label{fig:intro-diffusion}
    \end{subfigure}\hfill
    \begin{subfigure}{0.49\linewidth}
        \centering
        \includegraphics[width=\linewidth]{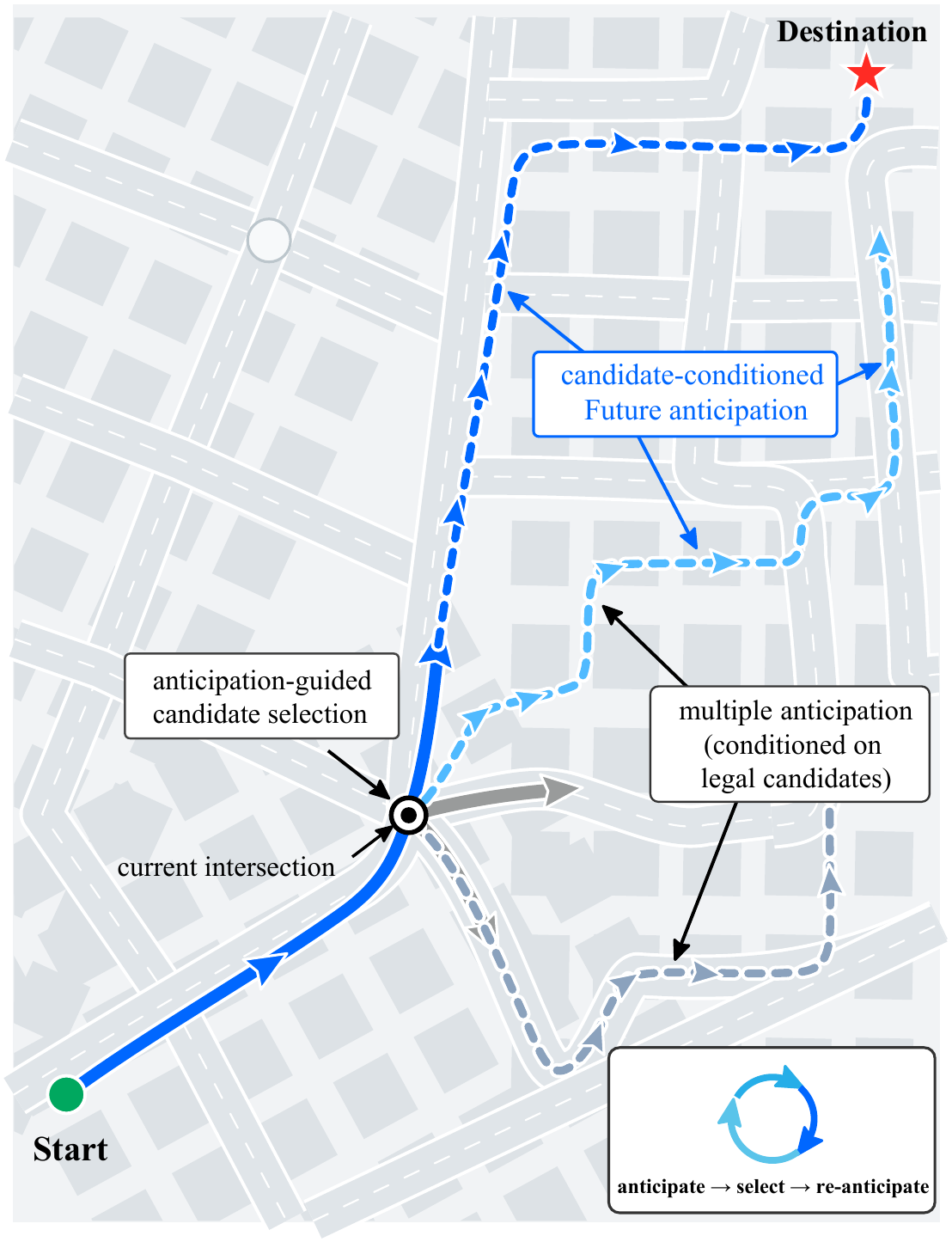}
        \caption{\textbf{\textsc{Foresight}}: Anticipatory route generation.}
        \label{fig:intro-foresight}
    \end{subfigure}
    \vspace{-0.05in}
    \caption{Motivated example for anticipatory generative modeling.}
    \label{fig:intro-motivation}
    \vspace{-0.25in}
\end{wrapfigure}
Generating human mobility data, particularly vehicle movement trajectories, plays a fundamental role in a wide spectrum of urban applications, including traffic management, navigation, and urban planning ~\citep{shi2024graphconstrained,li2026trajflow}. Within road networks, vehicle movement is constrained by the physical topology: movement along road segments is largely deterministic, whereas \emph{transitions at intersections are critical}—multiple outgoing paths may appear equally plausible given the traversed history, yet they can lead to substantially different downstream routes.

Conditioned on a given destination, a central challenge in road-network trajectory generation lies in modeling the next transition selection, especially at road intersections. In the literature, deep generative models such as diffusion-based methods are frequently employed~\citep{zhu2023difftraj, Wei2024diffrntraj,zhu2024controltraj,shi2024graphconstrained,li2026trajflow}. These models typically treat trajectories as holistic sequences of 2D locations and synthesize complete trajectories by denoising from random noise, often overlooking the discrete decision of next-transition selection at intersections. Consequently, they are prone to producing drift-off trajectories that deviate from the underlying road network, as illustrated in Fig.~\ref{fig:intro-motivation}\subref{fig:intro-diffusion}, which undermines their effectiveness for capturing real-world human mobility patterns. Regarding transition selection, classical search or dynamic programming, e.g., Dijkstra and A-star, attempt to find optimal solutions by exhaustively enumerating the action space, but they struggle to scale to large-scale road networks~\citep{dijkstra1959note,hart1968aformal}. Although reinforcement learning has been explored for action selection, the sparse reward signal associated with destination reachability makes it ill-conditioned for generating stepwise action sequences on road networks~\citep{bello2017neural,kool2018attention,sun2023hierarchical}. That is, route planning for road-network trajectories remains a largely open problem.

In this paper, we revisit trajectory generation on road networks, shifting from location simulation towards behavior modeling, and introduce a new graph-constrained generative paradigm, \emph{Anticipatory Generative Modeling}, for route planning. The core of this paradigm is \textbf{anticipation}, a provisional short-horizon continuation used to guide next-transition selection. A transition generator remains ambiguous when conditioned solely on the traversed history, since every legal transition on the road network may appear equally plausible. As illustrated in Fig.~\ref{fig:intro-motivation}\subref{fig:intro-foresight}, the situation changes when the model incorporates anticipation of how the future route may unfold toward the destination. Accordingly, we construct a 4-phase loop of “\underline{\emph{anticipate--select--update--re-anticipate}}” that enables longer-term foresight while continuing sequential generation along the historical trajectory.

Specifically, we present an anticipatory generation model named \textsc{Foresight} to produce plausible trajectories on road networks.  Given the destination $g$, we formalize \emph{Anticipatory Generative Modeling} (AGM) at the heart of our model, which first predicts a provisional, destination-conditioned continuation over a finite horizon, and then uses its representation to select the current transition:
\begin{equation}
    c_t,g \;\longrightarrow\; R_t=(S_{t+1},\ldots,S_{t+H})
    \;\longrightarrow\; q_t(\cdot),
\end{equation}
where $R_t$ is inferred from the available traversed history $c_t$. It thus serves as predictive context for the current selection, rather than being executed. After each transition, the history is updated and a new continuation is predicted. This loop preserves closed-loop generation while exposing downstream context. $H$ limits the explicitly represented future window, not the total trajectory length.
\textsc{Foresight} employs a Future Decoder to predict $R_t$, and a topology-constrained Transition Network to score and normalize legal outgoing candidates. A direction-aware fiber state encodes the geometry of each realized movement, while candidate-conditioned re-anticipation predicts the continuation induced by each hypothetical candidate after the current distribution is formed. The training objective couples the likelihood of the current legal transition with future-window prediction and candidate-conditioned future compatibility. During inference, an optional future-consistency score can further refine candidate selection. After each executed movement, \textsc{Foresight} recomputes the continuation from the updated history. The independent \texttt{STOP} event is admissible only at the destination, ensuring that validity and termination are governed by the same graph-constrained process.
Our key contributions are summarized as follows:
\begin{itemize}[leftmargin=1em]
\item \emph{Generative Route Planning.} We revisit trajectory generation through the lens of behavior modeling and fundamentally shift the focus from location simulation to route planning on road networks.
\item \emph{Anticipatory Generative Modeling.} Essentially different from diffusion models, we introduce a new graph-constrained generative paradigm, which resolves next transition selection by incorporating the anticipation of how the future route may unfold toward the destination.
\item \emph{Methodology Innovation.}  We purpose a novel \textsc{Foresight} model that formulates a probabilistic model with a direction-aware fiber for “\emph{anticipate–select–re-anticipate}” loop that enables longer-term foresight while continuing sequential generation along the historical trajectory.
\item \emph{Empirical Validation.} \textsc{Foresight} achieves state-of-the-art performance across various metrics, demonstrating its effectiveness in generating plausible trajectories.
\end{itemize}

\section{Related Work}
\label{sec:related-work}

\paragraph{Trajectory generation.}
Trajectory generation methods span autoregressive models that recursively predict subsequent locations or road segments and latent-variable or diffusion models that synthesize trajectories more globally. In mobility modeling, diffusion-based approaches such as DiffTraj generate complete GPS sequences~\citep{zhu2023difftraj}, while road-network methods increasingly incorporate topology into either step-wise or sequence-level generation. Autoregressive methods such as ProgRPGAN~\citep{fu2021progprgan}, Semantic Routing~\citep{zhaosemantic}, and HOSER~\citep{cao2025holistic} construct routes incrementally, conditioning each local choice on the destination, map semantics, or traversed history. Graph-Constrained Diffusion, Diff-RNTraj, and ControlTraj instead generate complete paths or trajectories while explicitly encoding road-network structure~\citep{shi2024graphconstrained,Wei2024diffrntraj,zhu2024controltraj}. These approaches therefore primarily model either the immediate next transition or the trajectory as a complete sequence.

\paragraph{Iterative replanning.}
Receding-horizon control offers a complementary closed-loop paradigm: the planner repeatedly predicts over a finite horizon, executes the first action, and replans from the resulting state. Related principles have been adopted in learned driving planners that reason over continuous trajectories, occupancy representations, or traffic-scene forecasts.~\citep{liu2026bridgedrive,zheng2025diffusionbased,huang2026occdriver,jiang2024Scene,wu2024smart}. Foresight brings this interface to a discrete road graph: each graph-legal transition is informed by an explicitly predicted, goal-conditioned future window, while a position- and direction-aware fiber state preserves local context. After execution, the anticipated continuation is retained when the transition is aligned and revised when the rollout deviates, coupling trajectory generation with closed-loop replanning.

\paragraph{Planning with future information.}
Classical search methods reason over candidate futures: Dijkstra~\citep{dijkstra1959note} accumulates path costs, A star~\citep{hart1968aformal} adds heuristic estimates, and Yen's algorithm enumerates alternative loopless paths~\citep{Yen1971find}. Learned search and routing methods incorporate longer-term consequences through neural heuristics, returns, value functions, or target-conditioned policies~\citep{RyoYonetani2021Path,jain2021,bello2017neural,kool2018attention,zeng2023target}. Sequence-modeling and diffusion-based approaches instead represent or generate imagined trajectories~\citep{janner2021offline,chen2021decision,wu2023mtm,Janner2022plan}. These works encode future information through search, value estimates, or trajectory prediction. Foresight explicitly represents a goal-conditioned short-horizon continuation and uses it to select the next legal graph transition.

\section{Preliminaries and Problem Formulation}
\label{sec:preliminaries}
\label{sec:problem-formulation}

\paragraph{Road-network trajectories.}
We represent the road network as a directed graph
\(\mathcal G=(\mathcal V,\mathcal E)\), where \(\mathcal V\) is the set of
road locations and \(\mathcal E\) contains directed connections. Let
\(X_t\in\mathcal V\) denote the location at step \(t\), and let
\(A_t=\psi(X_t,X_{t-1})\) denote the direction-aware attribute of the
incoming movement (with an initialization token at \(t=0\)). The corresponding
road state is \(S_t=(A_t,X_t)\), and the realized history before the next
decision is \(c_t=S_{0:t}\). A trajectory is conditioned on a destination
\(g\in\mathcal V\). Each movement follows the graph topology, with
\(X_{t+1}\in\mathcal N_{\mathcal G}^{+}(X_t)\), where
\(\mathcal N_{\mathcal G}^{+}(X_t)\) denotes the outgoing neighbors of
\(X_t\). For a neighbor \(u\in\mathcal N_{\mathcal G}^{+}(X_t)\), define the
complete movement candidate
$s_{t+1}^{(u)}=\bigl(\psi(u,X_t),u\bigr)$.
We denote by
$\mathcal U_t^{\mathrm{move}}
=\{s_{t+1}^{(u)}:u\in\mathcal N_{\mathcal G}^{+}(X_t)\}$
the set of legal movement candidates available at step \(t\). The complete
legal event set additionally includes the independent termination event only
at the destination:
\[
\mathcal U_t=
\begin{cases}
\mathcal U_t^{\mathrm{move}}, & X_t\neq g,\\
\mathcal U_t^{\mathrm{move}}\cup\{\mathrm{STOP}\}, & X_t=g.
\end{cases}
\]

\paragraph{Actions and stopping time.}
At step \(t\), the model selects a legal event \(A_{t+1}\) after observing
\(c_t\). A movement event is one of the complete-state candidates in
\(\mathcal U_t^{\mathrm{move}}\) and produces its associated location
\(X_{t+1}\); \(\mathrm{STOP}\) creates no new road state.
Trajectory termination is defined by the stopping time
\begin{equation}
\tau=\inf\left\{t\geq 0:X_t = g, A_{t+1}=\mathrm{STOP}\right\}.
\label{eq:stopping-time}
\end{equation}
Because \(\mathrm{STOP}\) is legal only at the destination,
\(A_{\tau+1}=\mathrm{STOP}\) implies \(X_\tau=g\). Reaching \(g\) alone does
not terminate generation; the model must explicitly select
\(\mathrm{STOP}\). An external transition budget may truncate generation
before this selection, which is distinct from model-selected termination.

\paragraph{Conditional trajectory generation.}
Given the traversed history \(c_t\) and destination \(g\), the task is to
generate a topologically valid continuation and terminate it at the
destination. A conventional sequential formulation selects a legal action
according to
\(p(A_{t+1}\mid c_t,g)\), appends the resulting road location after a movement,
and repeats until \(\mathrm{STOP}\) is selected. Such a formulation makes the
next decision directly from the observed history and destination. We instead
study how to inform this local decision with the trajectory continuation that
each candidate movement may induce. The conditional future of interest is
\(P(S_{t+1:\tau}\mid c_t,g)\), whose first event determines the next legal
transition and whose remaining states describe a possible downstream
continuation.

\begin{figure}[t]
    \centering
    \includegraphics[width=\linewidth]{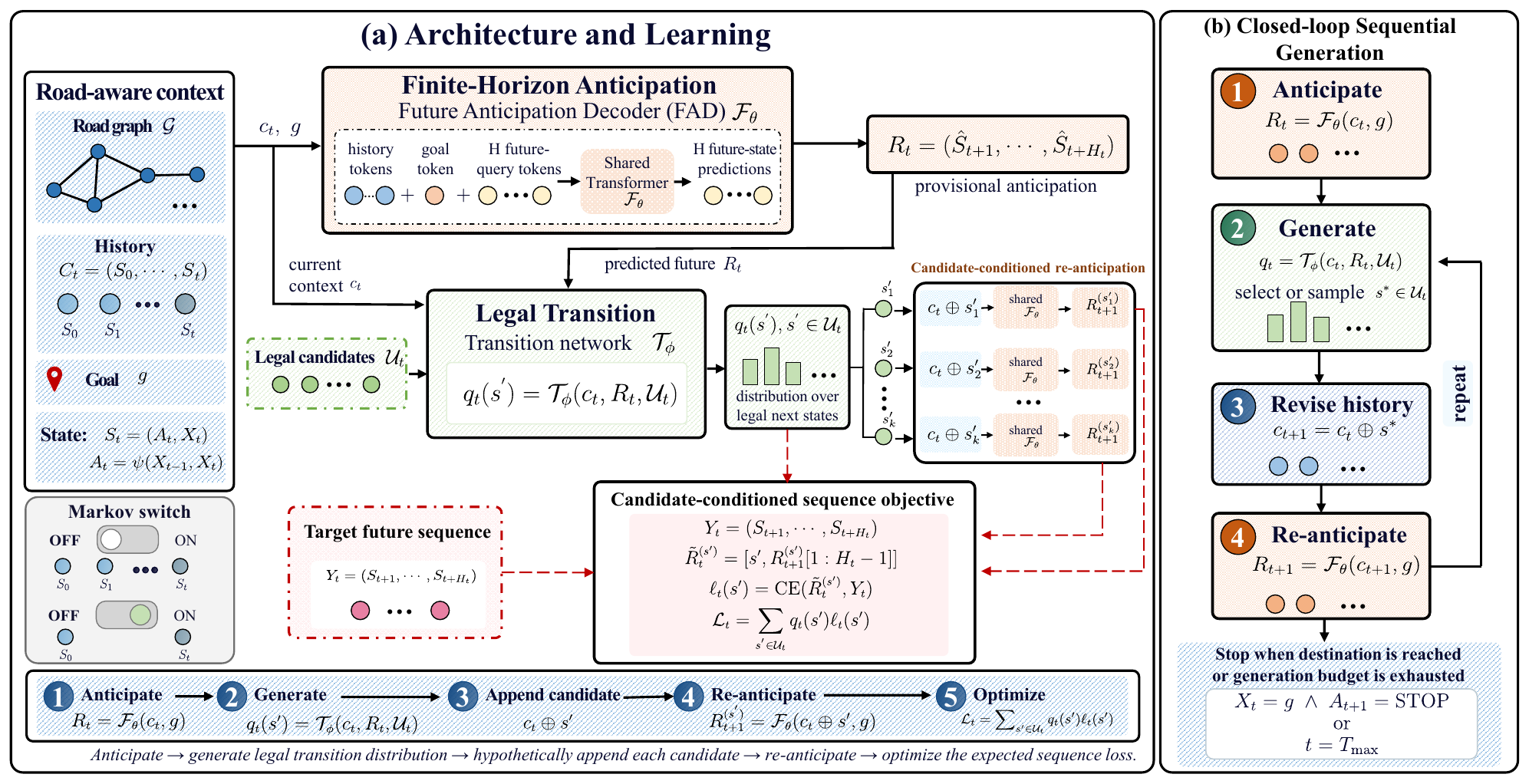}
    \caption{
    \textbf{Overview of \textsc{Foresight}.} Further architectural details are provided in Appendix~\ref{app:architecture}.
}
    \label{fig:architecture-overview}
\end{figure}

\section{Methodology}
\label{sec:remaining-method}
Distinguished from previous methods, we advocate a fundamental shift toward generative route planning on road networks. To resolve the  next-transition ambiguity that arises from relying solely on the traversed history, we introduce a new graph-constrained generative paradigm, Anticipatory Generative Modeling (AGM), which addresses this challenge by anticipating how the future route unfolds toward the destination. Accordingly, we propose an anticipatory model, \textsc{Foresight}, whose overall architecture is sketched in Fig.  \ref{fig:architecture-overview}. \emph{It recursively executes an “anticipate–select–update–re-anticipate” loop for stepwise routing on the road network.} Specifically, the Future Decoder first constructs an anticipation of a provisional short-horizon continuation. Then, the Transition Network selects the topologically valid outgoing direction based on its consistency with the anticipated future; the selected transition is appended to update the traversed sequence, and the model re-anticipates the trajectory suffix upon committing to the transition. This process repeats until \texttt{STOP} is chosen at the destination. Altogether, this endows the routing generator with longer-term foresight capacity while maintaining sequential generation along the historical trajectory.
\subsection{Anticipatory Generative Modeling}
\label{sec:anticipatory-generative-modeling}
We model goal-conditioned road trajectories as realizations of an underlying stochastic process. Given an origin \(x_0\) and destination \(g\), each training trajectory \(s_{0:\tau}^{(n)}\) is sampled from \(P_{\mathrm{data}}(S_{0:\tau}\mid X_0=x_0,g)\) and represents one feasible route between the two endpoints. Trajectories with the same endpoints may follow different paths, so an observed trajectory is one possible realization of the process rather than a unique target that all generated trajectories must match.

At an anchor indexed by \(t\), the traversed history \(c_t\) and destination \(g\) induce a conditional future process \(P(S_{t+1:\tau}\mid c_t,g)\). The suffix observed in a training example is one realization of this process: its first element specifies the next state, while the remainder records one possible downstream continuation. At inference, \(\textbf{\textsc{Foresight}}\) has access only to the traversed history and destination, from which it forms an anticipation of how the trajectory may continue.

A conventional stepwise generator directly models the next-state conditional \(P(S_{t+1}\mid c_t,g)\), corresponding to the one-step marginal of the full future process:
\[
P(s_{t+1}\mid c_t,g)
=
\sum_{s_{t+2:\tau}}
P(s_{t+1},s_{t+2:\tau}\mid c_t,g).
\]
Although longer-term regularities may still be learned from data, they are not explicitly represented as anticipations of downstream evolution that directly inform the current transition.

Anticipatory Generative Modeling (AGM) instead makes plausible downstream evolution explicit before selecting the current transition. Given \((c_t,g)\), \(\textbf{\textsc{Foresight}}\) anticipates how the trajectory may continue and uses this information to choose among topologically feasible transitions, without prescribing a unique future route.

Because the stopping time \(\tau\) is random, the future continuation \(S_{t+1:\tau}\) has variable length and becomes increasingly uncertain at longer horizons. Rather than representing the entire continuation before every transition, \(\textbf{\textsc{Foresight}}\) realizes AGM through a finite anticipation window, leaving more distant evolution to subsequent predictions.

\subsection{Finite-Horizon Anticipation}
\label{sec:finite-horizon-anticipation}
For a realized trajectory, each traversed prefix \(c_t\) defines an \emph{anchor} at step \(t\), from which the conditional future process extends to the random stopping time \(\tau\). Since \(\tau\) is unknown when the current decision is made, explicitly representing the full continuation would require modeling a variable-length predictive object. We instead restrict the explicitly represented future to a fixed horizon of \(H\) positions. Let \(Y_t\) denote the corresponding realized future window:
\begin{equation}
Y_t=S_{t+1:t+H},
\qquad
Y_t\sim P_{\mathrm{data}}(Y_t\mid c_t,g).
\label{eq:method-finite-anticipation}
\end{equation}
Here, \(Y_t\) provides the realized training target, whereas the model-predicted finite-horizon anticipation is denoted by \(\widehat R_t\). The horizon \(H\) bounds the amount of future context exposed to the current decision rather than the length of the generated trajectory. If \(\mathrm{STOP}\) occurs within the window, all subsequent positions are masked.

During training, the observed window \(Y_t\) supervises the predictive distributions represented by \(\widehat R_t\). During inference, \textbf{\textsc{Foresight}} predicts \(\widehat R_t\) from \((c_t,g)\) and uses its representation as downstream context when selecting the next legal transition.

The anticipation is revised after every realized movement:
\begin{equation}
c_t \longrightarrow \widehat R_t \longrightarrow S_{t+1}
\longrightarrow c_{t+1} \longrightarrow \widehat R_{t+1},
\qquad
\widehat R_{t+1}=\mathcal F_{\theta}(c_{t+1},g).
\label{eq:method-receding-anticipation}
\end{equation}
Thus, the \(H\)-step anticipation is not executed open-loop; it is recomputed from the updated history after each realized movement.

\subsection{\textsc{Foresight}: Neural Architecture} 
\label{sec:foresight-architecture}


We instantiate finite-horizon anticipation with \textbf{\textsc{Foresight}}, which couples a shared Future Anticipation Decoder with a topology-constrained Transition Network. Figure~\ref{fig:architecture-overview} summarizes the architecture and its step-wise generation process. Given the traversed history \(c_t\) and destination \(g\), the decoder predicts a finite-horizon anticipation, whose contextual representations are then used by the Transition Network to score the legal transitions from the current state:
\begin{equation}
\widehat R_t=\mathcal F_{\theta}(c_t,g),
\qquad
q_t=\mathcal T_{\phi}(c_t,g,\widehat R_t;\mathcal U_t).
\label{eq:method-foresight-overview}
\end{equation}
Here, \(\widehat R_t\) contains contextual representations for the next \(H\) positions together with the parameters of their associated categorical distributions, and \(q_t\) is the distribution over the legal candidate set \(\mathcal U_t\). We denote the categorical distribution at future position \(h\) by \(\widehat P_{t,h}\).

The Future Anticipation Decoder \(\mathcal F_{\theta}\) jointly encodes the traversed history, destination, and an ordered sequence of future queries with a shared Transformer. Its hidden states provide contextual representations for the next \(H\) positions, while prediction heads parameterize categorical distributions over the global predictive space. A structured visibility rule allows each future query to attend to the observed history, destination, and preceding query context, while preventing access to realized future states. The resulting anticipation therefore depends only on information available at the current anchor.

From the current road state \(x_t\), Foresight forms the movement candidate set \(\mathcal U_t^{\mathrm{move}}\) from the topologically valid outgoing transitions. The legal candidate set is
\[
\mathcal U_t
=
\begin{cases}
\mathcal U_t^{\mathrm{move}}\cup\{\mathrm{STOP}\}, & x_t=g,\\
\mathcal U_t^{\mathrm{move}}, & x_t\neq g,
\end{cases}
\]
so \(\mathrm{STOP}\) is available only at the destination. The Transition Network scores every legal candidate with the same compatibility function:
\begin{equation}
\operatorname{score}_t(s')
=
\operatorname{score}_{\phi}(c_t,g,\widehat R_t,s'),
\qquad
q_t(s')
=
\frac{\exp\,\operatorname{score}_t(s')}
{\sum_{\bar s\in\mathcal U_t}\exp\,\operatorname{score}_t(\bar s)},
\qquad
s'\in\mathcal U_t.
\label{eq:method-transition-score}
\end{equation}
Thus, movement candidates and \(\mathrm{STOP}\) are treated uniformly within the same topology-constrained distribution, while the anticipated future directly informs their scores.

After \(q_t\) has been formed, each movement candidate \(s'\in\mathcal U_t^{\mathrm{move}}\) can be appended hypothetically to the traversed history and re-evaluated by the same decoder:
\begin{equation}
c_t^{(s')}=c_t\oplus s',
\qquad
\widehat R_{t+1}^{(s')}=\mathcal F_{\theta}(c_t^{(s')},g),
\qquad
s'\in\mathcal U_t^{\mathrm{move}}.
\label{eq:method-candidate-continuation}
\end{equation}
These candidate-conditioned predictions represent the continuations induced by hypothetical legal movements. They are not separate observed targets; Section~\ref{sec:learning-objective} combines them with the single realized continuation to define the sequence learning objective.

\subsection{Learning Objective}
\label{sec:learning-objective}

For an anchor \(t\) on a sampled training trajectory, we extract an \(H\)-position future reference
\[
y_t=(y_{t,1},\ldots,y_{t,H}).
\]
This reference is one realization of the conditional future process given the traversed history and goal, rather than a unique target route. Its first entry \(y_{t,1}\) is the realized next event, while \(y_{t,2:H}\) records the subsequent continuation. If \(\mathrm{STOP}\) occurs within the horizon, its position is retained and all subsequent positions are masked; if \(y_{t,1}=\mathrm{STOP}\), only the first position contributes. The reference is used only for training supervision and is unavailable at inference.

To combine the current transition with subsequent future predictions in the same \(H\)-position objective, all positions are represented in a common prediction space. The Transition Network defines \(q_t\) over the local action set \(\mathcal U_t\), whereas future positions are predicted over the global space \(\mathcal Y=\mathcal V\cup\{\mathrm{STOP}\}\). We therefore map \(q_t\) to a global distribution \(\bar q_t\), assigning each movement probability to its resulting road state, assigning the STOP probability to \(\mathrm{STOP}\) when applicable, and setting all other entries to zero. Within the transition-coupled objective, \(\bar q_t\) provides the prediction for \(y_{t,1}\).

The Future Decoder is directly supervised by the masked future-window cross-entropy
\[
\mathcal L_{F,t}
=
-\sum_{h=1}^{H}m_{t,h}\log \widehat P_{t,h}(y_{t,h}).
\]
For each admissible movement candidate \(s'\in\mathcal U_t^{\mathrm{move}}\), we further use the hypothetical re-anticipation
\[
\widehat R_{t+1}^{(s')}
=
\mathcal F_{\theta}(c_t\oplus s',g)
\]
to obtain candidate-conditioned predictions for positions \(h=2,\ldots,H\). Since the training data provide no counterfactual continuation for unexecuted candidates, every candidate-conditioned prediction is evaluated against the same observed suffix \(y_{t,2:H}\). This suffix serves as a compatibility reference rather than a factual continuation conditioned on the hypothetical candidate. Let \(\widetilde P_{t,h}^{(s')}\) denote the corresponding global prediction at downstream position \(h\). We define
\[
\ell_t^{\mathrm{comp}}(s')
=
-\sum_{h=2}^{H}
m_{t,h}\log
\widetilde P_{t,h}^{(s')}(y_{t,h}).
\]

The anchor-level transition-coupled objective is
\begin{equation}
\mathcal L_t
=
-m_{t,1}\log\bar q_t(y_{t,1})
+
\sum_{s'\in\mathcal U_t^{\mathrm{move}}}
q_t(s')\,\ell_t^{\mathrm{comp}}(s').
\label{eq:method-anchor-objective}
\end{equation}
The first term supervises the realized next event, while the second evaluates the expected downstream compatibility of candidate movements under \(q_t\). The complete anchor loss is
\[
\mathcal J_t
=
\mathcal L_{F,t}+\mathcal L_t.
\]
Candidate-conditioned re-anticipation is computed only after \(q_t\) is formed and therefore affects the transition distribution only through the training objective. Training averages \(\mathcal J_t\) over sampled trajectories and temporal anchors.

\subsection{Sequential Generation}
\label{sec:sequential-generation}

At generation step \(t\), given the realized history \(c_t\) and destination \(g\), \textsc{Foresight} first predicts the current finite-horizon anticipation and uses it to define a base distribution over the legal action set:
\begin{equation}
\widehat R_t
=
\mathcal F_{\theta}(c_t,g),
\qquad
q_t
=
\mathcal T_{\phi}(c_t,g,\widehat R_t;\mathcal U_t).
\label{eq:method-inference-selection}
\end{equation}
Generation uses only the realized history, the destination, and model-predicted anticipations; the future reference used for training is never available. By default, the next action is selected directly from \(q_t\). Optionally, \textsc{Foresight} can further rerank legal movements according to the consistency between the current anticipation and the continuation predicted after each hypothetical candidate.

For a movement candidate \(s'\in\mathcal U_t^{\mathrm{move}}\), let
\[
\widehat R_{t+1}^{(s')}
=
\mathcal F_{\theta}(c_t\oplus s',g)
\]
denote its candidate-conditioned re-anticipation, and let
\(\widehat P_{t,h}\) and
\(\widehat P_{t+1,h}^{(s')}\)
denote the categorical distributions associated with position \(h\) of
\(\widehat R_t\) and \(\widehat R_{t+1}^{(s')}\), respectively. Since advancing by one movement shifts the prediction horizon by one position, future consistency is measured over their overlapping predictions:
\begin{equation}
E_{\mathrm{fut}}(s')
=
\frac{1}{H-1}
\sum_{h=1}^{H-1}
D_{\mathrm{JS}}
\left(
\widehat P_{t,h+1}
\,\Vert\,
\widehat P_{t+1,h}^{(s')}
\right),
\qquad
s'\in\mathcal U_t^{\mathrm{move}},
\label{eq:method-inference-js}
\end{equation}
where \(D_{\mathrm{JS}}\) is the Jensen--Shannon divergence. We set
\(E_{\mathrm{fut}}(\mathrm{STOP})=0\).
The resulting selection energy and distribution are
\begin{equation}
E_t(s')
=
-\log q_t(s')
+
wE_{\mathrm{fut}}(s'),
\quad
\pi_t(s')
=
\frac{\exp[-E_t(s')]}
{\sum_{u\in\mathcal U_t}\exp[-E_t(u)]},
\quad
\hat a_{t+1}
=
\operatorname{Select}(\pi_t;\mathcal U_t).
\label{eq:method-inference-reranking}
\end{equation}
When \(w=0\), \(\pi_t=q_t\), recovering selection from the base transition distribution. Candidate-conditioned branches used in reranking remain hypothetical: they are evaluated only to score the current decision, and only the selected action \(\hat a_{t+1}\) is executed and appended to the realized history.

If \(\hat a_{t+1}\in\mathcal U_t^{\mathrm{move}}\), executing the selected movement determines the next road position. Let \(\operatorname{dest}(\hat a_{t+1})\) denote the destination road position of the selected legal movement. Consistent with the state definition \(S_t=(A_t,X_t)\), the realized state and updated history are
\begin{equation}
x_{t+1}
=
\operatorname{dest}(\hat a_{t+1}),
\qquad
s_{t+1}
=
(\hat a_{t+1},x_{t+1}),
\qquad
c_{t+1}
=
c_t\oplus s_{t+1}.
\label{eq:method-inference-update}
\end{equation}
\textsc{Foresight} then recomputes the anticipation from the updated realized history:
\begin{equation}
\widehat R_{t+1}
=
\mathcal F_{\theta}(c_{t+1},g).
\label{eq:method-inference-reanticipation}
\end{equation}
The anticipation therefore informs the current transition without prescribing an open-loop route. Each executed movement changes the realized history and induces a new prediction, yielding a closed-loop anticipate--select--update--re-anticipate process.

If \(\hat a_{t+1}=\mathrm{STOP}\), generation terminates immediately and no subsequent movement state is appended. Reaching \(x_t=g\) only makes \(\mathrm{STOP}\) admissible; termination still requires selecting it. A maximum generation budget provides an external evaluation-time truncation: reaching the budget before \(\mathrm{STOP}\) is recorded as a timeout rather than as stochastic termination.

\section{Experiments}
\label{sec:experiments}

We study whether Foresight can reproduce both the spatial distribution of held-out mobility and the route structure of individual trips. The experiments are organized around four questions: how the model compares with existing generators, whether its samples occupy space like the test set, what its generated routes look like under matched conditions, and which parts of the anticipatory state are responsible for the gains.

\begin{table}[t]
\centering
\caption{Overall performance across cities. \textbf{D-JSD/R-JSD denote Distance/Radius JSD and are reported in units of $10^{-2}$.} HD is Hausdorff distance and EDR is edit distance on real sequences. Cells report OD-bootstrap means under the common evaluation protocol (see Appendix~\ref{app:full-results}). Lower is better.}
\label{tab:experiment-main}
\fontsize{9pt}{9pt}\selectfont\setlength{\tabcolsep}{1pt}
\resizebox{0.96\linewidth}{!}{%
\begin{tabular*}{\linewidth}{@{\extracolsep{\fill}}l*{12}{c}}
\toprule
& \multicolumn{4}{c}{Porto} & \multicolumn{4}{c}{San Francisco} & \multicolumn{4}{c}{Beijing}\\
\cmidrule(lr){2-5}\cmidrule(lr){6-9}\cmidrule(lr){10-13}
Method & D-JSD & R-JSD & HD & EDR & D-JSD & R-JSD & HD & EDR & D-JSD & R-JSD & HD & EDR\\
\midrule
Dijkstra & 2.47 & 1.31 & 0.481 & 0.358 & 2.18 & 1.29 & \underline{0.399} & \underline{0.305} & 2.64 & 1.56 & 0.449 & 0.341\\
Yen-$k$ & 2.36 & 1.29 & 0.503 & 0.386 & 2.34 & 1.39 & 0.497 & 0.377 & 2.78 & 1.67 & 0.552 & 0.414\\
Markov & 2.61 & 1.46 & 0.613 & 0.462 & 2.39 & 1.42 & 0.602 & 0.458 & 2.88 & 1.72 & 0.681 & 0.503\\
SemRoute & 1.68 & 1.07 & 0.514 & 0.384 & \underline{1.69} & 0.91 & 0.481 & 0.363 & 2.16 & 1.19 & 0.548 & 0.414\\
\midrule
HOSER & 1.57 & 0.96 & 0.431 & 0.317 & 1.82 & 0.98 & 0.412 & 0.306 & 2.04 & 1.09 & 0.463 & 0.351\\
TS-TrajGen & 3.24 & 1.83 & 0.641 & 0.481 & 3.06 & 1.86 & 0.608 & 0.462 & 3.42 & 2.16 & 0.684 & 0.503\\
GDP & 2.43 & 1.57 & 0.531 & 0.397 & 2.31 & 1.31 & 0.501 & 0.376 & 2.76 & 1.68 & 0.553 & 0.417\\
Diff-RNTraj & 2.34 & 1.39 & 0.476 & 0.357 & 2.19 & 1.28 & 0.417 & 0.319 & 2.63 & 1.54 & 0.468 & 0.352\\
DiffTraj & 2.83 & 1.76 & 0.637 & 0.479 & 2.57 & 1.61 & 0.617 & 0.468 & 3.01 & 1.98 & 0.692 & 0.504\\
TrajFlow & 1.72 & 1.08 & 0.536 & 0.401 & 1.71 & 0.88 & 0.472 & 0.354 & 2.23 & 1.21 & 0.546 & 0.411\\
Seed & 1.79 & 1.14 & 0.528 & 0.398 & 1.86 & 1.04 & 0.506 & 0.382 & 2.28 & 1.29 & 0.559 & 0.419\\
\midrule
\textbf{\textsc{Foresight}-F} & \underline{1.53} & \underline{0.79} & \underline{0.401} & \underline{0.312} & \textbf{1.39} & \underline{0.71} & \textbf{0.386} & \textbf{0.298} & \textbf{1.68} & \textbf{0.88} & \textbf{0.418} & \textbf{0.326}\\
\textbf{\textsc{Foresight}-M} & \textbf{1.01} & \textbf{0.71} & \textbf{0.375} & \textbf{0.285} & 2.68 & \textbf{0.65} & 0.542 & 0.415 & \underline{1.87} & \underline{1.01} & \underline{0.438} & \underline{0.344}\\
\bottomrule
\end{tabular*}%
}

\end{table}
\FloatBarrier

\begin{table}[!ht]
\centering
\caption{Ablation results for Porto and San Francisco. \textbf{JSD values are scaled by $10^{-2}$.} }
\label{tab:experiment-ablation}
\small
\setlength{\tabcolsep}{3pt}
\resizebox{0.96\linewidth}{!}{%
\begin{tabular*}{\linewidth}{@{\extracolsep{\fill}}l*{4}{c}@{\hspace{10pt}}*{4}{c}@{}}
\toprule
& \multicolumn{4}{c}{Porto} & \multicolumn{4}{c}{San Francisco}\\
\cmidrule(lr){2-5}\cmidrule(lr){6-9}
Variant & D-JSD & R-JSD & HD & EDR & D-JSD & R-JSD & HD & EDR\\
\midrule
\textsc{Foresight}-F & \underline{1.53} & \underline{0.79} & \underline{0.401} & \underline{0.312} & \textbf{1.39} & \underline{0.71} & \textbf{0.386} & \textbf{0.298}\\
\textsc{Foresight}-M & \textbf{1.01} & \textbf{0.71} & \textbf{0.375} & \textbf{0.285} & 2.68 & \textbf{0.65} & 0.542 & 0.415\\
\textit{w/o }$\mathcal L_{F,t}$ & 2.25 & 1.35 & 0.500 & 0.390 & 2.10 & 1.25 & 0.480 & 0.380\\
\textit{w/o }$\ell_t^{\mathrm{comp}}$ & 2.55 & 1.55 & 0.550 & 0.420 & 2.40 & 1.45 & 0.530 & 0.410\\
\textit{w/o }$E_{\mathrm{fut}}$ & 2.05 & 1.20 & 0.460 & 0.360 & 1.95 & 1.15 & 0.450 & 0.350\\
\bottomrule
\end{tabular*}%
}
\end{table}
\FloatBarrier

\paragraph{Experimental setup.}
We evaluate goal-conditioned generation on the Porto, Beijing, and San Francisco datasets with the official HOSER graphs and splits.\footnote{The HOSER dataset is available at \url{https://huggingface.co/datasets/caoji2001/HOSER-dataset}.} All methods use matched test OD conditions, replicate counts, and generation budgets; we report Distance/Radius JSD, Hausdorff distance, and EDR (lower is better). Full protocol, training, and metric details appear in Appendices~\ref{app:experiments}.

\paragraph{Baselines.}
Baselines span the Markov reference, Dijkstra and Yen's $k$-shortest-path search~\citep{dijkstra1959note,Yen1971find}, Semantic Routing and HOSER~\citep{zhaosemantic,cao2025holistic}, and learned generators: TS-TrajGen~\citep{ContinuousTrajectoryGene}, GDP~\citep{shi2024graphconstrained}, Diff-RNTraj~\citep{Wei2024diffrntraj}, DiffTraj~\citep{zhu2023difftraj}, TrajFlow~\citep{li2026trajflow}, and Seed~\citep{rao2025seed}. Adaptation and implementation details are in Appendix~\ref{app:experiments}; all methods use the matched evaluation protocol above.
\paragraph{Overall performance.}
Table~\ref{tab:experiment-main} reports distribution-level and route-level fidelity under the common evaluation protocol. Distance JSD and Radius JSD are reported in units of $10^{-2}$; complete results with bootstrap standard deviations, DTW, and completion diagnostics are provided in Appendix~\ref{app:full-results}. The two Foresight variants show complementary behavior: the full-history variant is strongest on San Francisco and Beijing, whereas the Markov-history variant performs best on Porto and remains competitive elsewhere. Completion diagnostics use the same test conditions and retain failed rollouts in their denominators.
\paragraph{Spatial distributions and route examples.}
The quantitative comparison is complemented by spatial and route views under matched held-out conditions; the complete spatial and route panels are in Appendix~\ref{app:full-visuals}.

\begin{figure}[htb]
\centering
\begin{tabular}{@{}c@{\hspace{0.04\linewidth}}c@{}}
    \includegraphics[width=0.44\linewidth,height=1.45in,keepaspectratio]{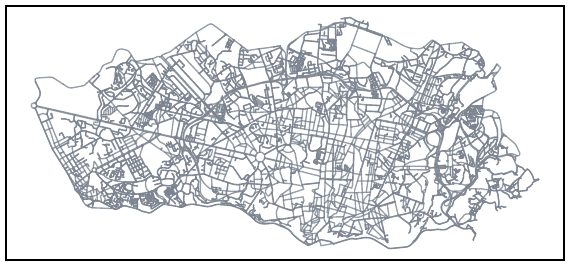} &
    \includegraphics[width=0.44\linewidth,height=1.45in,keepaspectratio]{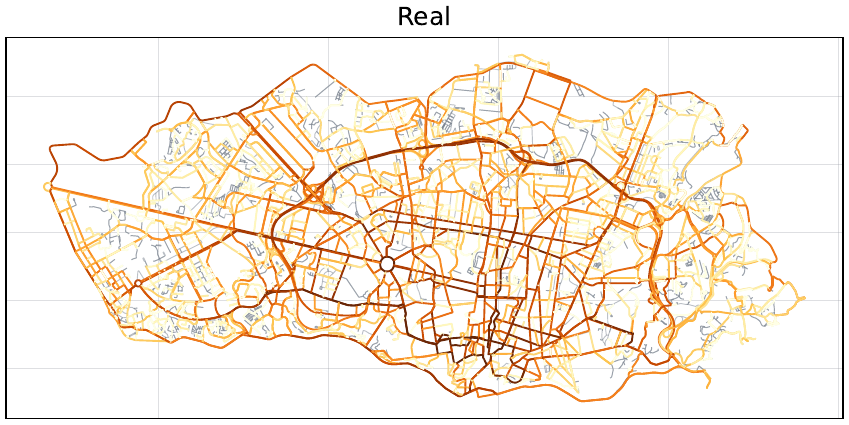}\\[-1pt]
    {\footnotesize Road network} & {\footnotesize Test distribution}\\[5pt]
    \includegraphics[width=0.44\linewidth,height=1.45in,keepaspectratio]{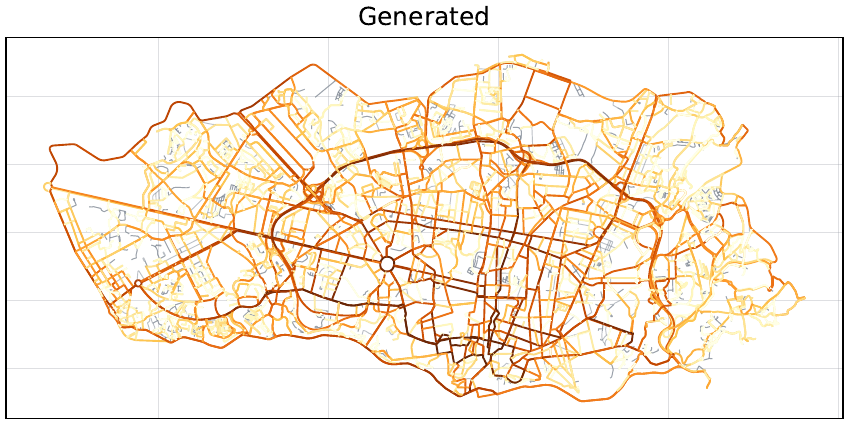} &
    \includegraphics[width=0.44\linewidth,height=1.45in,keepaspectratio]{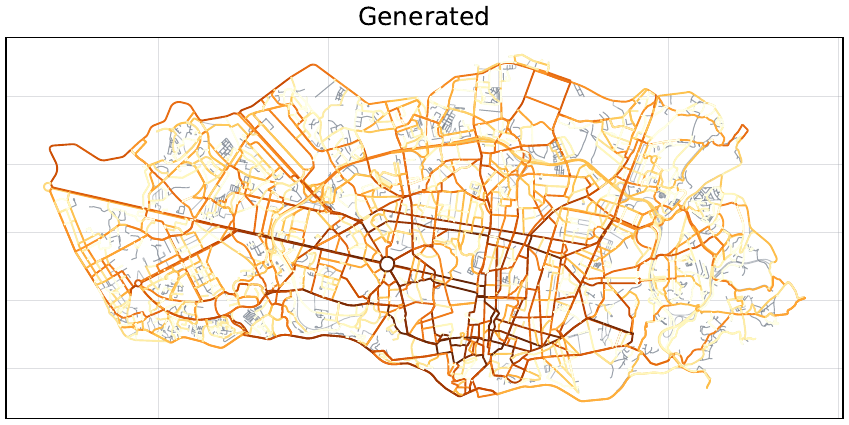}\\[-1pt]
    {\footnotesize Generated: Foresight} & {\footnotesize Generated: Dijkstra}
\end{tabular}
\caption{Porto spatial comparison. Top: road network and test-set trajectory distribution. Bottom: trajectory distributions generated by Foresight and Dijkstra.}
\label{fig:experiment-spatial}
\end{figure}

\paragraph{Ablation of anticipatory modeling.}
We ablate complete trajectory history, future-window supervision $\mathcal L_F$, candidate-conditioned continuation loss $\ell_t^{\mathrm{con}}$, and inference-time future energy $E_{\mathrm{fut}}$. As shown in Table~\ref{tab:experiment-ablation}, candidate-conditioned continuation contributes the most, followed by future-window supervision, while future-energy reranking provides a smaller but consistent improvement. Full results, including dataset-specific history effects, standard deviations, DTW, and Beijing experiments, are provided in Appendix~\ref{app:full-results}.
\begin{figure}[htb]
\centering
\begin{tabular}{@{}c@{\hspace{0.02\linewidth}}c@{\hspace{0.02\linewidth}}c@{}}
    \includegraphics[width=0.32\linewidth,height=1.45in,keepaspectratio]{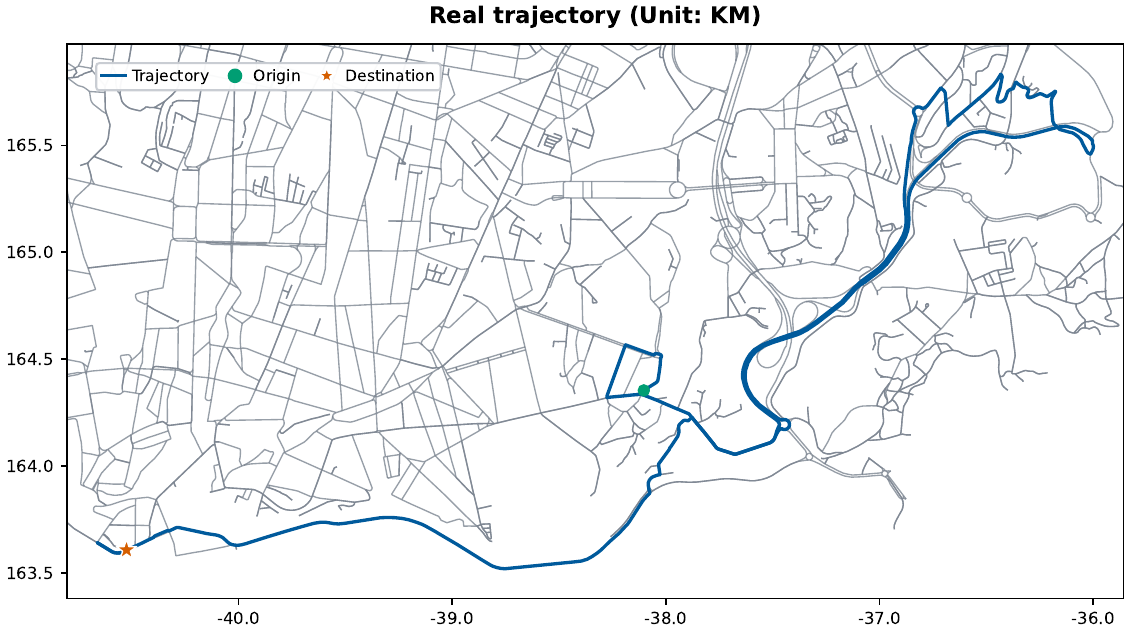} &
    \includegraphics[width=0.32\linewidth,height=1.45in,keepaspectratio]{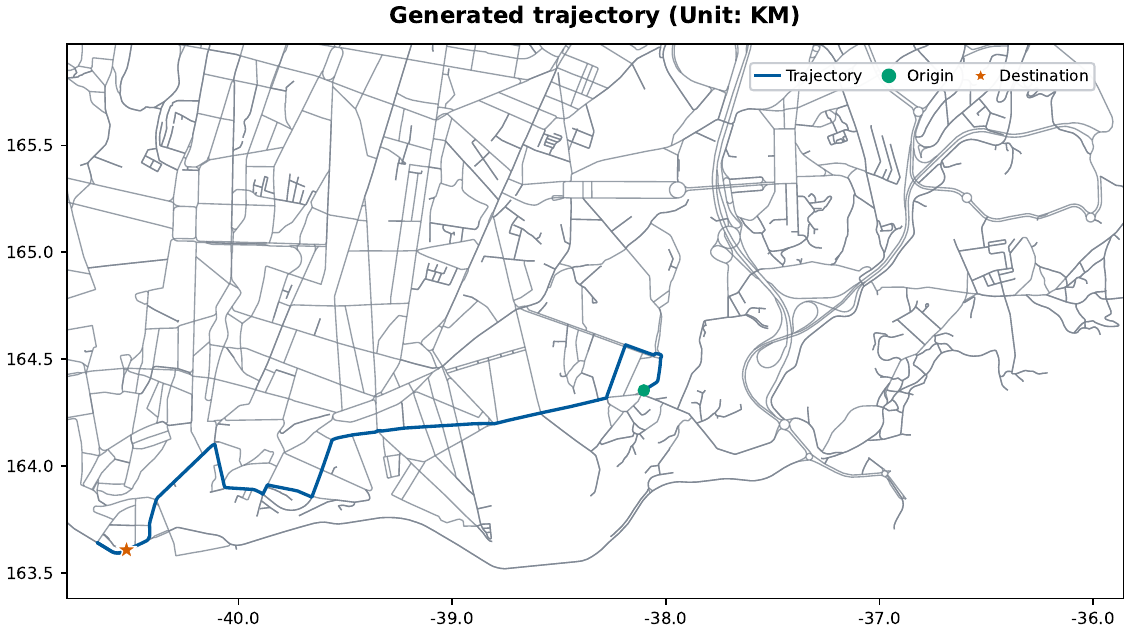} &
    \includegraphics[width=0.32\linewidth,height=1.45in,keepaspectratio]{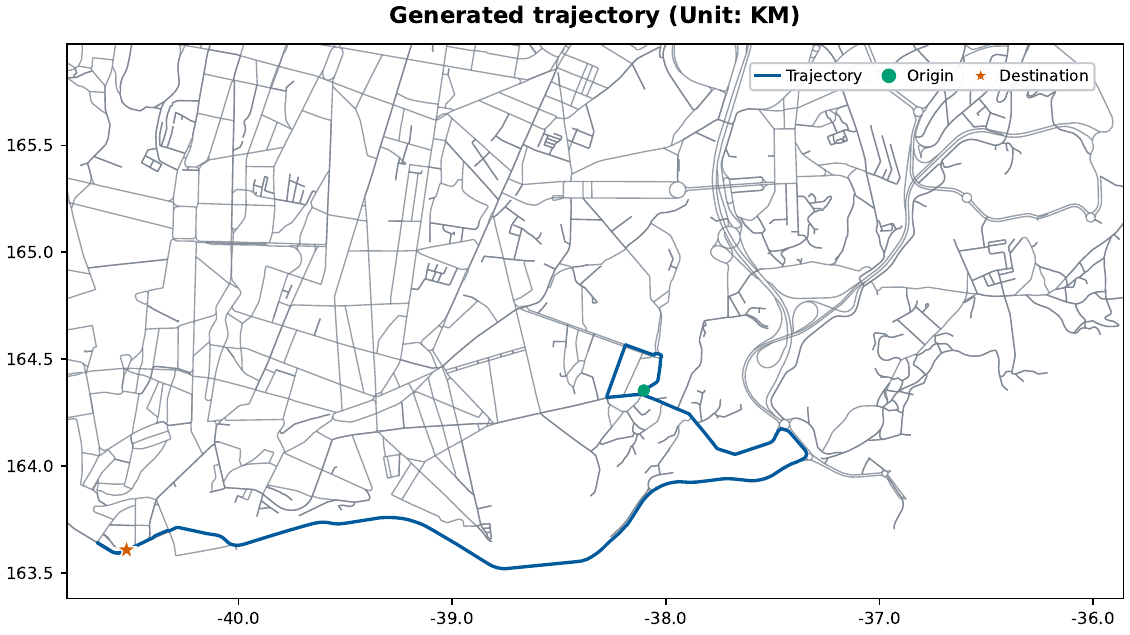}\\[4pt]
    \includegraphics[width=0.32\linewidth,height=1.45in,keepaspectratio]{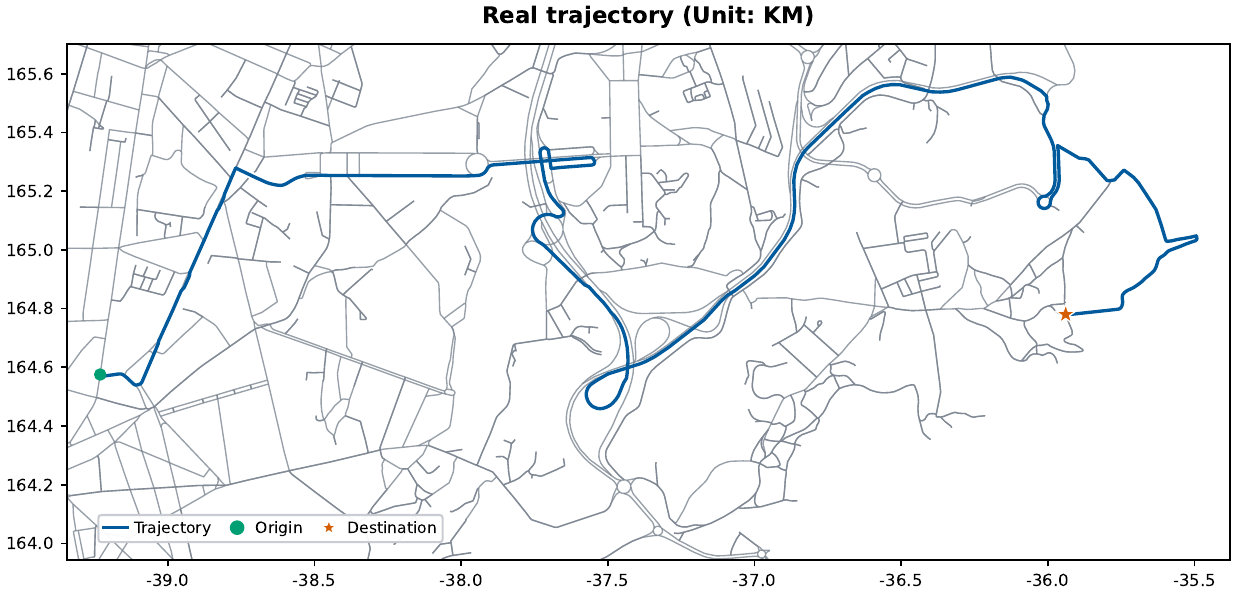} &
    \includegraphics[width=0.32\linewidth,height=1.45in,keepaspectratio]{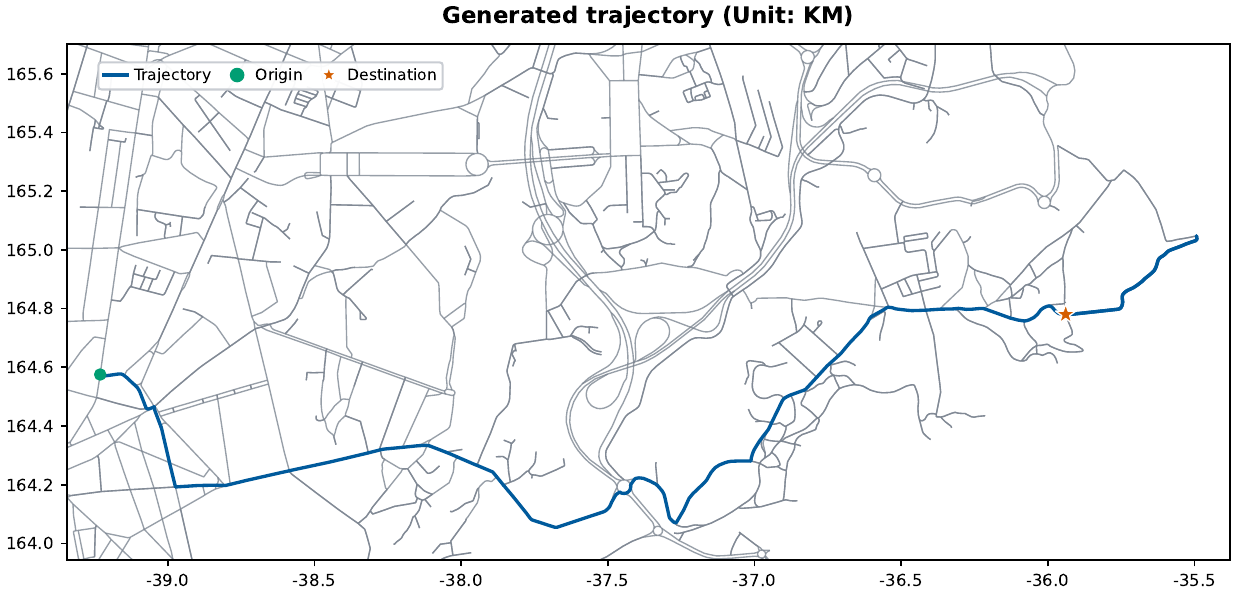} &
    \includegraphics[width=0.32\linewidth,height=1.45in,keepaspectratio]{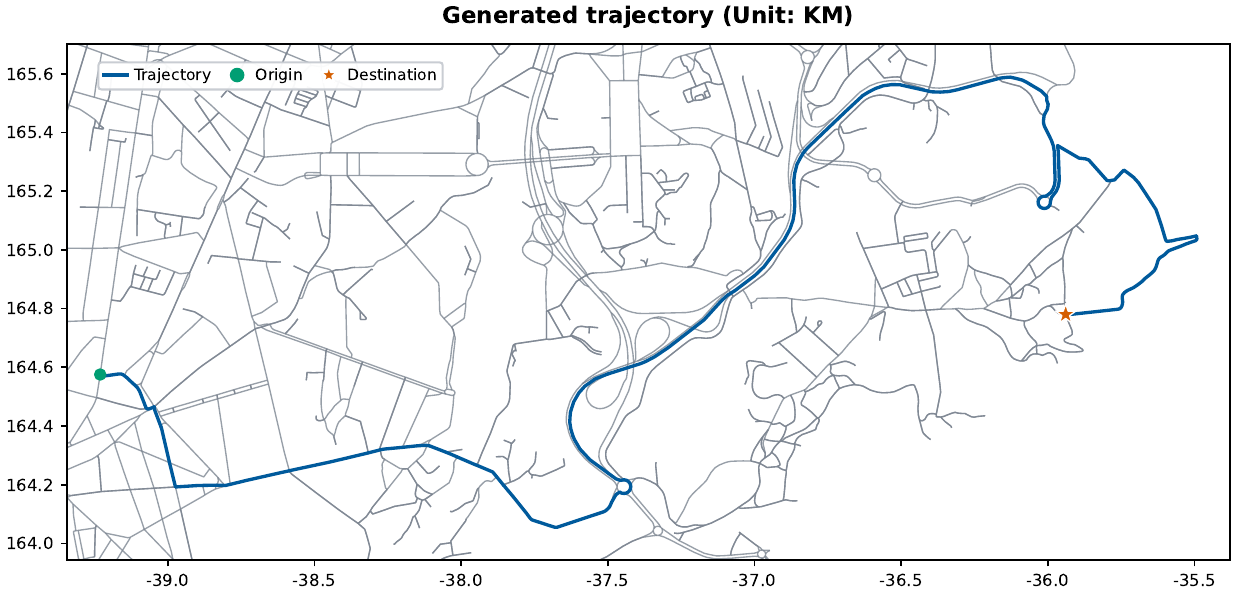}\\[-1pt]
    {\footnotesize Real Trajectory} & {\footnotesize Dijkstra} & {\footnotesize Foresight}
\end{tabular}
\caption{Matched Porto route comparison for two fixed test OD conditions.}
\label{fig:experiment-paths}
\end{figure}

\section{Conclusion}
\label{sec:conclusion}
This work revisits trajectory generation through the lens of generative route planning on road networks. To this end, we introduce a new graph-constrained generative paradigm, Anticipatory Generative Modeling, which resolves the ambiguity in next-transition selection by anticipating how the future route unfolds toward the destination. Specifically, we present an anticipatory model, \textsc{Foresight}. At each step, it (i) constructs an \emph{anticipation} of a provisional short-horizon continuation, (ii) \emph{selects} among topologically valid transitions via a direction-aware fiber, (iii) \emph{updates} the traversed sequence through the decoder, and (iv) \emph{re-anticipates} the trajectory suffix upon committing to the transition. Experiments on real-world datasets show that \textsc{Foresight} achieves superior performance in both trajectory distribution matching and route fidelity under origin–destination constraints. These results suggest that the proposed anticipation mechanism helps disambiguate locally plausible transitions while preserving topological validity throughout sequential generation.
\section*{AI Use Statement}

AI tools were used solely for language polishing and formatting adjustments during manuscript preparation. They were not used to develop the research ideas or methodology, conduct experiments, analyze experimental results, or determine the scientific claims and conclusions of this work. The authors reviewed and take responsibility for all content in the manuscript.

\section*{Reproducibility Statement}

We document the inputs and procedures needed to reproduce the stated experiments. We use the public HOSER road graphs and official train/validation/test splits for Porto, Beijing, and San Francisco (\url{https://huggingface.co/datasets/caoji2001/HOSER-dataset}). Preprocessing reads \texttt{roadmap.geo}, \texttt{roadmap.rel}, and the supplied trajectory files; it validates road identifiers, graph-valid transitions, timestamp consistency, and finite attributes. Continuous-feature statistics are fitted on training data and reused for validation, test, and generation. Trajectories are not re-split, and test trajectories do not fit preprocessing statistics or model parameters. Each run is bound to a versioned data/configuration manifest, a validation-selected checkpoint, and a fixed test-condition and replicate list.

The method appendices fix the model interfaces and training boundaries. Shared Base/Fiber encoders construct history-consistent node--fiber states and enumerate only graph-legal successors. A structured Future Transformer uses a fixed $H=64$ query window, global node-ID targets, and an independent \texttt{STOP} event; its visibility mask excludes future labels from the context. The Transition Network builds a candidate-independent \texttt{q\_trans} readout, then scores complete legal candidate states with a projected dot product and positive temperature. Full-history and Markov variants share this interface; the latter retains only $[S_0,S_t]$ and the goal in both modules. One joint optimizer transaction combines masked future event cross-entropy, current-event NLL, and expected future cross-entropy over candidate branches. Future targets are stop-gradient, Transition inputs are detached, and discrete actions are detached before history append. The reported training configuration uses AdamW for 60,000 updates with fixed validation checkpoint selection; each run records scale dimensions, parameter counts, normalization settings, precision, clipping, and random states. The appendix lists the five scale presets, their widths, depths, heads, horizons, and city-dependent vocabulary parameter counts for direct checkpoint comparison.

Generation uses the frozen model in evaluation/no-gradient mode. At every decision it rebuilds the fixed future window from the realized history, passes the downstream future representation to Transition, scores legal outgoing graph edges, and admits \texttt{STOP} only at the goal. Reaching the goal alone does not terminate a rollout. The reported protocol uses 5,000 fixed test conditions, and a maximum of 256 transitions per attempt; greedy or sampling selection and candidate-conditioned future-consistency reranking are run-level fields. Every attempt is retained, including timeouts, invalid transitions, and numerical failures, with condition/OD identifiers, seed, node sequence, stop reason, and checkpoint/configuration references. Baselines share the graph, test conditions, replicate policy, and failure-accounting rule; their adapters, decoding/search settings, and budgets are recorded separately.

Evaluation is reproducible from saved generation records and the evaluation configuration. The primary metrics are Distance JSD, Radius JSD, Hausdorff distance, DTW, and EDR; undefined metrics stay unavailable rather than becoming zeros. Completion statistics use the complete attempted corpus as their denominator. Exact-OD bootstrap is primary and uses 1,000 resamples with bootstrap seed 0; condition-bootstrap and HOSER-style file-level aggregation are secondary checks. The experimental appendix specifies supports, binning, matching rules, and Python/PyTorch-style pseudocode for training, inference, and evaluation. 
Cells without measured values identify missing runs or inputs and are not interpreted as results.

\bibliographystyle{plainnat}
\bibliography{references}
\clearpage
\appendix
\section{Notation}
\label{app:notation}

Table~\ref{tab:notation} summarizes the symbols used in the problem
formulation, model, and training objective. We distinguish road locations,
direction-aware states, legal events, and the global output alphabet. The
training target $y_t$ is a realized reference window, whereas
$\widehat R_t$ is the model-predicted anticipation.

\begin{table}[!ht]
\centering
\caption{Core notation used in the paper.}
\label{tab:notation}
\small
\begin{tabular}{@{}p{0.30\linewidth}p{0.62\linewidth}@{}}
\toprule
Symbol & Meaning \\
\midrule
\multicolumn{2}{@{}l}{\emph{Trajectory, state, and legal events}} \\
$\mathcal G=(\mathcal V,\mathcal E)$ & Directed road graph. \\
$X_t\in\mathcal V$ & Location at step $t$. \\
$X_{0:T}$ & Location trajectory; $T$ is an index (and may be the endpoint of a realized trajectory). \\
$g\in\mathcal V$ & Destination location. \\
$A_t=\psi(X_t,X_{t-1})$ & Direction/fiber attribute induced by the incoming movement; $A_0$ is the initialization token. \\
$S_t=(A_t,X_t)$ & Direction-aware road state. \\
$c_t=S_{0:t}$ & Realized direction-aware state history before the next decision. \\
$\mathcal N_{\mathcal G}^+(X_t)$ & Outgoing neighbors of $X_t$. \\
$s'(u)=(\psi(u,X_t),u)$ & Complete direction-aware movement candidate for neighbor $u$. \\
$\mathcal U_t^{\mathrm{move}}$ & Legal movement candidates, one for each outgoing neighbor. \\
$\mathcal U_t$ & Legal event support; it adds $\mathrm{STOP}$ only when $X_t=g$. \\
$\tau$ & Model stopping time: the first step at which the selected event is $\mathrm{STOP}$; this is legal only when $X_t=g$. \\
$\oplus$ & History append operation, e.g., $c_{t+1}=c_t\oplus S_{t+1}$. \\
\midrule
\multicolumn{2}{@{}l}{\emph{Anticipation, targets, and candidate branches}} \\
$\widehat R_t=\mathcal F_{\theta}(c_t,g)$ & Model-predicted finite-horizon anticipation with $H$ positions (written as $R_t$ in the probabilistic appendix). \\
$H$ & Anticipation and supervision-window length. \\
$y_t=(y_{t,1},\ldots,y_{t,H})$ & Realized training target window over locations and STOP; it retains STOP and uses PAD after STOP. \\
$\mathcal Y=\mathcal V\cup\{\mathrm{STOP}\}$ & Global output alphabet. \\
$\mathrm{PAD}$ & Storage-only padding symbol; it is not in $\mathcal Y$. \\
$m_{t,h}\in\{0,1\}$ & Mask: valid through STOP, then zero. \\
$\widehat P_{t,h}$ & Decoder categorical distribution over $\mathcal Y$ at anticipation position $h$. \\
$\mathcal L_{F,t}$ & Masked future-window cross-entropy for the decoder. \\
$\mathcal F_{\theta}$ & Future Anticipation Decoder mapping $(c_t,g)$ to $\widehat R_t$. \\
$\mathcal T_{\phi}$ & Transition Network mapping $(c_t,g,\widehat R_t,\mathcal U_t)$ to legal-event scores/probabilities. \\
$q_t(e)$ & Probability of legal event $e\in\mathcal U_t$. \\
$\bar q_t$ & Sparse $q_t$ aligned to $\mathcal Y$ via $\iota$. \\
$\iota$ & Event-to-output alignment: $\iota(s'(u))=u$ and $\iota(\mathrm{STOP})=\mathrm{STOP}$. \\
$c_t^{(s')}=c_t\oplus s'$ & History after appending candidate $s'$. \\
$\widehat R_{t+1}^{(s')}$ & Candidate-conditioned re-anticipation after appending movement $s'$. \\
\midrule
\multicolumn{2}{@{}l}{\emph{Learning and model parameters}} \\
$\widetilde P_{t,h}^{(s')}$ & Global distribution aligned at position $h$: $\bar q_t$ at $h=1$ and the candidate-conditioned decoder distribution afterward. \\
$\ell_t^{\mathrm{comp}}(s')$ & Masked suffix compatibility cross-entropy for movement candidate $s'$. \\
$\mathcal L_t$ & Transition-coupled anchor objective: current-event NLL plus the $q_t$-weighted candidate compatibility loss. \\
$\mathcal J_t$ & Complete anchor loss, $\mathcal L_{F,t}+\mathcal L_t$. \\
$E_{\mathrm{fut}}(s')$ & Optional future-consistency energy for movement candidate $s'$. \\
$\pi_t$ & Optional reranked legal-event distribution used for selection. \\
$\theta,\phi$ & Future Decoder and Transition Network parameters, respectively. \\
$\mathbf S$ & Random-length stopped state trajectory, $S_{0:\tau}$. \\
\bottomrule
\end{tabular}
\end{table}

The directional attribute is determined by the incoming movement, whereas the
destination condition is determined by node identity. The realized target
$y_t$ and mask $m_{t,h}$ are used only for training; at inference,
$\widehat R_t$ is predicted from $(c_t,g)$ and only the selected event is executed.
STOP is an event in $\mathcal U_t$ only at the destination and does not
correspond to a successor state.

\section{Probabilistic Foundation of Foresight}
\label{app:theory}

This appendix gives a probabilistic interpretation of Foresight that is consistent with Section~\ref{sec:remaining-method}. The underlying object is a random-length trajectory on a directed road network. A training trajectory is one realization of this process, and a future window extracted from it is one sampled realization at a temporal anchor. In contrast, $R_t$ denotes a model-predicted anticipation computed from the history and goal. Keeping these objects distinct is essential when hypothetical candidate-conditioned continuations are evaluated.

\subsection{Random-Length Road-Network Trajectory Process}
\label{app:theory-process}

Let $\mathcal G=(\mathcal V,\mathcal E)$ be a directed road network and let $X_t\in\mathcal V$ denote the road state at step $t$. For an ordinary movement, the direction-aware action is induced by two consecutive road states,
\begin{equation}
 A_t=\psi(X_t,X_{t-1})\in\mathcal A_{\mathrm{move}},\qquad
 S_t=(A_t,X_t),\quad t\geq 1,
 \label{eq:app-theory-state}
\end{equation}
where $A_0=\mathrm{BOS}$ and $S_0=(\mathrm{BOS},X_0)$. Thus $S_t$ is induced by the road-state process; $A_t$ is not an additional independent latent process. The realized history before the next decision is
\begin{equation}
 c_t=S_{0:t},\qquad g\in\mathcal V,
 \label{eq:app-theory-history}
\end{equation}
where $g$ is the destination and $X_0$ is the origin encoded by $S_0$.

The action alphabet includes an explicit terminal event,
\begin{equation}
 \mathcal A^+=\mathcal A_{\mathrm{move}}\cup\{\mathrm{STOP}\}.
 \label{eq:app-theory-action-alphabet}
\end{equation}
The stopping time is defined by that event alone:
\begin{equation}
 \tau=\inf\{t\geq 0:A_{t+1}=\mathrm{STOP}\}.
 \label{eq:app-theory-stopping-time}
\end{equation}
At $X_t=g$, STOP becomes admissible, but reaching the destination does not terminate the process automatically. If $A_{t+1}\in\mathcal A_{\mathrm{move}}$, a new road state is realized and $A_{t+1}=\psi(X_{t+1},X_t)$; if $A_{t+1}=\mathrm{STOP}$, no ordinary $X_{t+1}$ or $S_{t+1}$ is generated. With the decision filtration
\begin{equation}
 \mathcal F_t=\sigma(S_0,A_1,S_1,\ldots,S_t,A_{t+1}),
 \label{eq:app-theory-filtration}
\end{equation}
$\tau$ is a stopping time. The complete stopped trajectory is
\begin{equation}
 \mathbf S=S_{0:\tau}=(S_0,\ldots,S_\tau),
 \label{eq:app-theory-stopped-trajectory}
\end{equation}
whose length is random. At the data level,
\begin{equation}
 \mathbf S^{(n)}\sim P_{\mathrm{data}}(\mathbf S\mid X_0,g).
 \label{eq:app-theory-data-sample}
\end{equation}
A training trajectory is one realization of this goal-conditioned process, not a deterministic specification of a unique route between its endpoints. The conditional distribution is learned from regularities across trajectory samples and temporal anchors; a single realization does not identify the full multimodal distribution.

\subsection{Goal-Conditioned Future Process and Anticipatory Modeling}
\label{app:theory-future}

At an anchor $t\leq\tau$, the complete future is the variable-length suffix $S_{t+1:\tau}$ with conditional distribution
\begin{equation}
 P(S_{t+1:\tau}\mid c_t,g).
 \label{eq:app-theory-full-future}
\end{equation}
The usual next-step conditional is its one-step marginal,
\begin{equation}
 P(S_{t+1}\mid c_t,g)=\sum_{S_{t+2:\tau}}P(S_{t+1:\tau}\mid c_t,g).
 \label{eq:app-theory-next-marginal}
\end{equation}
The summation represents marginalization over downstream continuations. Long-range regularities may therefore be encoded implicitly, but the future is not an explicit context for the current decision. Anticipatory Generative Modeling (AGM) makes such context explicit by predicting a finite future representation before selecting the current legal transition.

For a sampled trajectory $\mathbf S^{(n)}$, the observed suffix $S_{t+1:\tau^{(n)}}^{(n)}$ is one realization of the conditional future for $(c_t^{(n)},g)$. It is a supervision reference, not a unique route and not an inference input. The learned computation is
\begin{equation}
 R_t=\mathcal F_{\theta}(c_t,g),\qquad
 q_t=\mathcal T_{\phi}(c_t,g,R_t;\mathcal U_t).
 \label{eq:app-theory-model-dependency}
\end{equation}
These arrows describe computation and do not assert a conditional-independence statement such as $S_{t+1}\perp(c_t,g)\mid R_t$.

\subsection{Finite-Horizon Anticipation}
\label{app:theory-finite-horizon}

The full future has random length, whereas the model exposes a fixed number $H$ of future positions at each decision. Let
\begin{equation}
 \mathcal Y=\mathcal V\cup\{\mathrm{STOP}\},\qquad
 y_t=(y_{t,1},\ldots,y_{t,H}),\quad y_{t,h}\in\mathcal Y\cup\{\mathrm{PAD}\}.
 \label{eq:app-theory-future-alphabet}
\end{equation}
For a sampled trajectory and anchor $t$, define the STOP-aware reference by
\begin{equation}
 y_{t,h}=\begin{cases}
 X_{t+h},&1\leq h\leq\tau-t,\\
 \mathrm{STOP},&h=\tau-t+1,\\
 \mathrm{PAD},&h>\tau-t+1,
 \end{cases}\qquad
 m_{t,h}=\mathbf 1\{h\leq\tau-t+1\}.
 \label{eq:app-theory-window-target}
\end{equation}
The STOP position is supervised, while positions after STOP are masked. PAD is only a storage symbol and is not a decoder output.

The model-predicted anticipation is
\begin{equation}
 R_t=\mathcal F_{\theta}(c_t,g),
 \label{eq:app-theory-finite-anticipation}
\end{equation}
whose $H$ prediction positions provide downstream context for the Transition Network. Thus $H$ is the explicit anticipation horizon, not the length of a generated trajectory, and fixing $H$ does not impose conditional independence on more distant states. The sampled reference $y_t$ is used only during training.

The anticipation is provisional and is recomputed after each realized movement:
\begin{equation}
 R_t\longrightarrow s_{t+1}\longrightarrow c_{t+1}\longrightarrow R_{t+1},\qquad
 c_{t+1}=c_t\oplus s_{t+1},\quad R_{t+1}=\mathcal F_{\theta}(c_{t+1},g).
 \label{eq:app-theory-receding}
\end{equation}

\subsection{Legal Transition, Candidate-Conditioned Prediction, and Learning}
\label{app:theory-candidate-future}

Let $\mathcal N_G^+(X_t)$ denote the outgoing neighbors of $X_t$. The legal movement candidates are
\begin{equation}
 \mathcal U_t^{\mathrm{move}}=\{s'(u):u\in\mathcal N_G^+(X_t)\},\qquad
 s'(u)=\bigl(\psi(u,X_t),u\bigr),
 \label{eq:app-theory-movement-support}
\end{equation}
and the current event support is
\begin{equation}
 \mathcal U_t=\begin{cases}
 \mathcal U_t^{\mathrm{move}},&X_t\neq g,\\
 \mathcal U_t^{\mathrm{move}}\cup\{\mathrm{STOP}\},&X_t=g.
 \end{cases}
 \label{eq:app-theory-legal-support}
\end{equation}
The transition distribution is normalized only on this topology-constrained support:
\begin{equation}
 q_t=\mathcal T_{\phi}(c_t,g,R_t;\mathcal U_t),\qquad
 \sum_{e\in\mathcal U_t}q_t(e)=1.
 \label{eq:app-theory-transition-distribution}
\end{equation}

Future positions use the global alphabet $\mathcal Y$, whereas current events use $\mathcal U_t$. With $\iota(s'(u))=u$ and $\iota(\mathrm{STOP})=\mathrm{STOP}$, align the current distribution by
\begin{equation}
 \bar q_t(y)=\sum_{\substack{e\in\mathcal U_t\\\iota(e)=y}}q_t(e),\qquad y\in\mathcal Y.
 \label{eq:app-theory-event-alignment}
\end{equation}
This is a probability-space alignment, not a second classifier.

After $q_t$ has been formed, each hypothetical movement candidate defines a candidate-conditioned history and re-anticipation:
\begin{equation}
 c_t^{(s')}=c_t\oplus s',\qquad
 R_{t+1}^{(s')}=\mathcal F_{\theta}(c_t\oplus s',g),\quad s'\in\mathcal U_t^{\mathrm{move}}.
 \label{eq:app-theory-candidate-continuation}
\end{equation}
The order is
\begin{equation}
 R_t\longrightarrow q_t\longrightarrow c_t\oplus s'\longrightarrow R_{t+1}^{(s')}.
 \label{eq:app-theory-candidate-order}
\end{equation}
$R_{t+1}^{(s')}$ is a model prediction under a hypothetical history, not an observed counterfactual continuation. STOP has no ordinary continuation branch.

Let $p_{t+1,j}^{(s')}$ be the decoder distribution at downstream position $j$. For alignment with positions $2,\ldots,H$, define
\begin{equation}
 \widetilde P_{t,1}^{(s')}=\bar q_t,\qquad
 \widetilde P_{t,h}^{(s')}=p_{t+1,h-1}^{(s')},\quad 2\leq h\leq H.
 \label{eq:app-theory-aligned-predictions}
\end{equation}
The sampled-future compatibility cost is
\begin{equation}
 \ell_t^{\mathrm{comp}}(s')=-\sum_{h=2}^{H}m_{t,h}\log\widetilde P_{t,h}^{(s')}(y_{t,h}).
 \label{eq:app-theory-compatibility}
\end{equation}
The same sampled suffix is a compatibility reference for all hypothetical branches. For an unexecuted candidate, $y_{t,2:H}$ is generally not a sample from $P_{\mathrm{data}}(\cdot\mid c_t\oplus s',g)$; the data contain no factual future for every counterfactual movement. The unified anchor criterion is
\begin{equation}
 \mathcal L_t=-m_{t,1}\log\bar q_t(y_{t,1})+\sum_{s'\in\mathcal U_t^{\mathrm{move}}}q_t(s')\ell_t^{\mathrm{comp}}(s').
 \label{eq:app-theory-unified-objective}
\end{equation}
The first term supervises the observed current event, including STOP. The second couples transition probabilities to sampled-future compatibility over hypothetical movement branches. It is one unified criterion rather than separately weighted transition, future, and rollout losses. Candidate re-anticipation is not a forward input to $q_t$, but its compatibility cost affects learning through the objective.

For valid trajectory-anchor pairs $\mathcal I$, the dataset objective is
\begin{equation}
 \widehat{\mathcal R}(\Theta)=\frac{1}{|\mathcal I|}\sum_{(n,t)\in\mathcal I}\mathcal L_t^{(n)},\qquad
 \mathcal R(\Theta)=\mathbb E_{\mathbf S\sim P_{\mathrm{data}}}\left[\sum_{t\leq\tau}\mathcal L_t(\mathbf S)\right].
 \label{eq:app-theory-empirical-risk}
\end{equation}
The empirical average is a Monte Carlo estimate of the expected value of the defined training criterion. This interpretation does not turn an unobserved branch into an observed counterfactual sample, nor does it claim that $\ell_t^{\mathrm{comp}}(s')$ is an unbiased estimator of a true counterfactual conditional likelihood.

\subsection{Sequential Generation as a Stochastic Process}
\label{app:theory-sequential}

At inference, only the realized history and destination are available:
\begin{equation}
 R_t=\mathcal F_{\theta}(c_t,g),\qquad
 q_t=\mathcal T_{\phi}(c_t,g,R_t;\mathcal U_t),\qquad
 \hat a_{t+1}=\operatorname{Select}(q_t;\mathcal U_t).
 \label{eq:app-theory-generation-step}
\end{equation}
The sampled training window $y_t$ is never provided. If $\hat a_{t+1}=s'\in\mathcal U_t^{\mathrm{move}}$, then
\begin{equation}
 s_{t+1}=s',\qquad c_{t+1}=c_t\oplus s_{t+1},\qquad R_{t+1}=\mathcal F_{\theta}(c_{t+1},g).
 \label{eq:app-theory-sequential-movement}
\end{equation}
If $\hat a_{t+1}=\mathrm{STOP}$, generation terminates with $\tau=t$ and no ordinary state is appended. A maximum generation budget may truncate a run before STOP, but it does not redefine $\tau$.

For stochastic action selection, the probability of a stopped realization is schematically
\begin{equation}
 P_\Theta(S_{1:\tau},A_{\tau+1}=\mathrm{STOP}\mid S_0,g)=\left[\prod_{t=0}^{\tau-1}q_t(S_{t+1})\right]q_\tau(\mathrm{STOP}),
 \label{eq:app-theory-stopped-probability}
\end{equation}
where each $q_t$ is recomputed from the current realized history. Training may evaluate all hypothetical movement branches for compatibility; inference realizes exactly one legal event and updates only that branch. The resulting closed-loop process is
\begin{equation}
\begin{aligned}
 &(X_{t-1},X_t)\to A_t=\psi(X_t,X_{t-1}),\quad (A_t,X_t)\to S_t;\\
 &(c_t,g)\to R_t\to q_t\to\hat a_{t+1};\\
 &\text{movement: }\hat a_{t+1}\to s_{t+1}\to c_{t+1}\to R_{t+1};\\
 &\text{STOP: }\hat a_{t+1}=\mathrm{STOP}\to\tau;\\
 &\text{training branch: }q_t\to s'\to c_t\oplus s'\to R_{t+1}^{(s')}\dashrightarrow y_t.
\end{aligned}
\label{eq:app-theory-branch-graph}
\end{equation}
The dashed link denotes training supervision only. It does not make $y_t$ a causal input to $q_t$ or an observed label for a hypothetical branch.

\section{Architecture and Inference Details}
\label{app:model-details}\label{app:architecture}

This appendix fixes the architectural and inference interfaces used in Section~\ref{sec:remaining-method}. The model uses a shared node--fiber representation, a single structured Future Anticipation Decoder, and a topology-constrained Transition Network. Throughout, $c_t=S_{0:t}$ denotes the realized state history, $S_t=(A_t,X_t)$ is the current direction-aware state, $g$ is the destination, $R_t=\mathcal F_{\theta}(c_t,g)$ is the predicted finite-horizon anticipation, and $q_t$ is the distribution over the legal event set $\mathcal U_t$. Uppercase block names such as $C$, $F$, $G$, and $T$ are reserved for the token blocks in the visibility mask. The learning criterion is defined in Section~\ref{sec:learning-objective}; this appendix specifies its tensor interfaces and the closed-loop inference procedure.

\subsection{Fiber-Aware State and Candidate Encoding}
\label{app:state-encoding}

The shared node encoder maps node attributes to $z_x=E_\theta(a_x)\in\mathbb R^{d_X}$. It is used for history nodes, the destination, and candidate nodes. Node IDs index the graph and the global vocabulary; they are not additional inputs to the node-attribute encoder. A visited state is represented causally as
\begin{equation}
    A_0=0,\qquad A_t=\psi(X_t,X_{t-1}),\quad t\ge 1,
    \qquad S_t=(A_t,X_t).
\end{equation}
Thus the current state contains the incoming transition that has already occurred. For a legal outgoing node $u\in\mathcal N_{\mathcal G}^+(X_t)$, the online candidate is
\begin{equation}
    A_{t+1}^{(u)}=\psi(u,X_t),\qquad
    s_{t+1}^{(u)}=(A_{t+1}^{(u)},u).
\end{equation}
The movement support used below is therefore
\begin{equation}
    \mathcal U_t^{\mathrm{move}}
    =\{s_{t+1}^{(u)}:u\in\mathcal N_{\mathcal G}^{+}(X_t)\}.
    \label{eq:app-movement-support}
\end{equation}
If this candidate is executed, its fiber becomes the incoming fiber of the next history state. Reaching the destination does not replace that fiber by a zero vector. A zero-fiber destination token is used only for destination conditioning. STOP is an independent event rather than a graph node, so it does not pass through either encoder.

The complete state embedding is formed before the Transformer input projection:
\begin{equation}
    \operatorname{Tok}(S_t)=W_S[z_{X_t}\Vert A_t]+e_{\mathrm{time}}(t)+e_{\mathrm{history}}.
    \label{eq:app-state-token}
\end{equation}
The destination-conditioning token uses $[z_g\Vert 0]$ and a destination role embedding. This token does not overwrite the incoming fiber of an actually visited goal state.

\subsection{Future Anticipation Decoder}
\label{app:future}

The Future Anticipation Decoder is one structured Transformer with token layout
\begin{equation}
 \mathcal X_t=[\operatorname{Tok}(c_t),Q_t^F,\operatorname{Tok}(g),Q_t^G,e_{\mathrm{STOP}}],
\end{equation}
where $Q_t^F$ contains $H$ ordered future queries and $Q_t^G$ contains learned global slots. Query $h$ carries the absolute time $t+h$ and the projected destination condition,
\begin{equation}
 q^F_{t,h}=e_{\mathrm{time}}(t+h)+W_S[z_g\Vert 0],
 \qquad h=1,\ldots,H.
\end{equation}
The global-slot position identifies a slot index, not a trajectory time.

\begin{figure}[t]
    \centering
    \includegraphics[width=\linewidth]{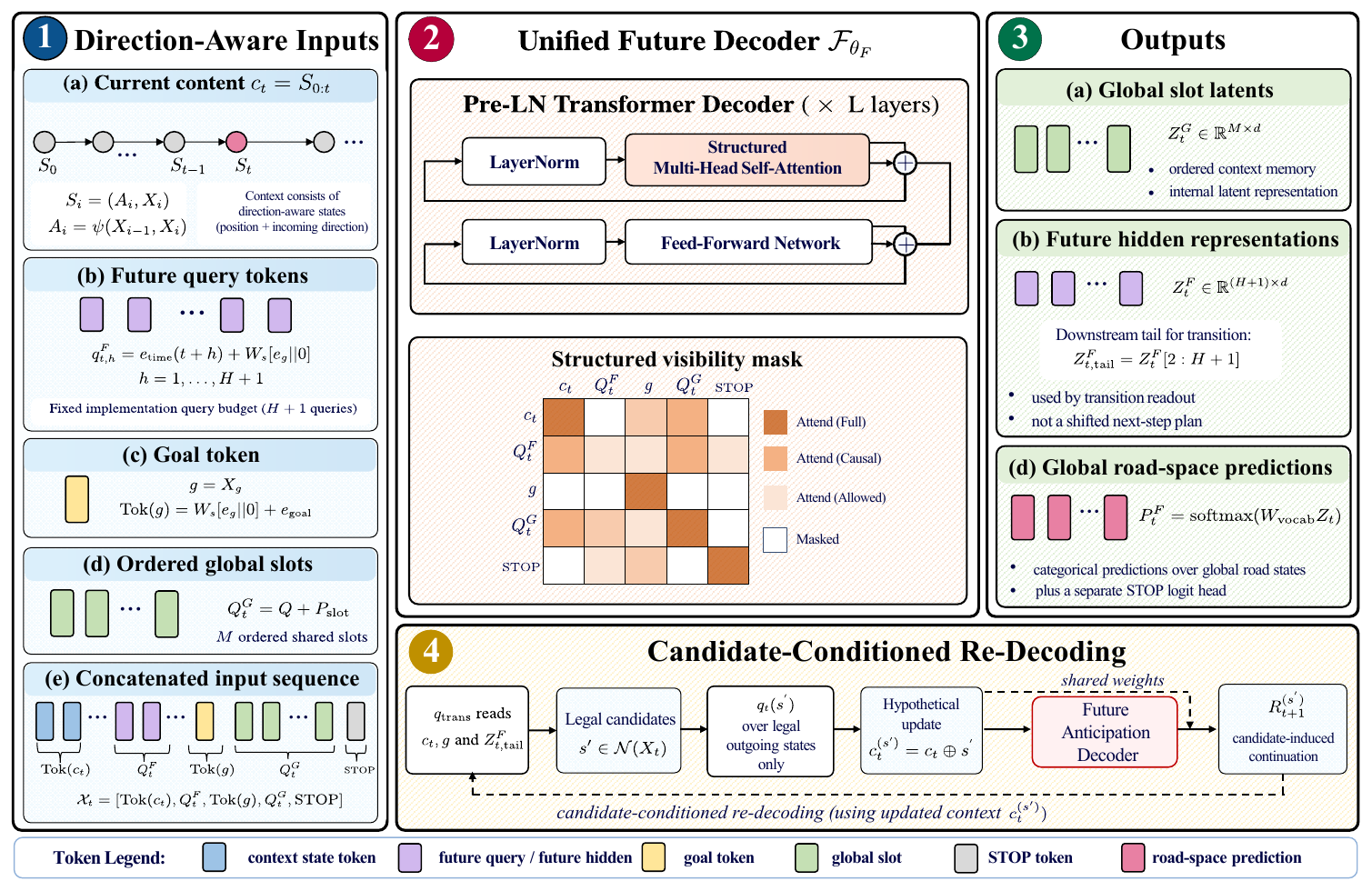}
    \caption{Future Anticipation Decoder. The structured Transformer maps the causal history $c_t$ and destination $g$ to global-slot latents, future hidden representations, and global road-space predictions. The downstream hidden tail is passed to the Transition Network; candidate-conditioned re-decoding reuses the same decoder after a hypothetical legal movement.}
    \label{fig:app-future-decoder}
\end{figure}

The structured visibility mask is applied at every Transformer layer. Let $C$, $F$, $g$, $G$, and $T$ denote history, future-query, destination, global-slot, and STOP blocks. The block-level visibility contract is:

\begin{table}[t]
\centering
\small
\caption{Visibility mask for the Future Anticipation Decoder. Padding masks additionally remove invalid rows and keys.}
\label{tab:future-visibility}
\begin{tabular}{c|ccccc}
\toprule
row $\backslash$ key & $C$ & $F$ & $g$ & $G$ & $T$\\
\midrule
$C$ & prefix & no & no & no & no\\
$F_h$ & yes & $F_{\le h}$ & yes & yes & yes\\
$g$ & no & no & self & no & no\\
$G$ & yes & no & yes & all $G$ & no\\
$T$ & no & no & no & no & self\\
\bottomrule
\end{tabular}
\end{table}

The structured visibility mask is applied at every Transformer layer. History tokens attend only within the current anchor's observed prefix. A future query attends to valid history, the destination, all valid global slots, and itself or earlier future queries. Destination and global slots do not read future queries, and later future queries cannot influence an earlier query. Different anchors never attend to one another. This mask cannot be replaced by an ordinary causal mask over the concatenated sequence.

\texttt{query\_valid} describes which query tokens are constructed and may be read by the network. At inference the requested fixed window supplies all $H$ valid queries, independent of the unknown realized suffix length. \texttt{target\_valid} is separate: it masks only the observed target sequence in training and offline comparison. The Future Anticipation Decoder therefore never receives the true suffix length, a realized future token, or a target mask as a conditioning feature.

The decoder interface is
\begin{equation}
    (Z_t^G,Z_t^F,L_t)=\mathcal F_{\theta}(c_t,g),
    \qquad Z_t^F\in\mathbb R^{H\times d},\quad
    L_t\in\mathbb R^{H\times|\mathcal Y|}.
    \label{eq:app-future-interface}
\end{equation}
The fixed query layout is constructed internally from $t$ and $H$. Each future position returns a hidden state and a categorical logit vector over $\mathcal Y$. The exported anticipation in the notation of Section~\ref{sec:remaining-method} is
\begin{equation}
    R_t=(Z_t^F,L_t).
    \label{eq:app-anticipation-output}
\end{equation}
The first hidden state and logit vector describe the immediate next event. The Transition Network consumes only the downstream hidden tail $Z_t^F[2{:}H]$ (equivalently, the future tensor with its first query removed) and its \texttt{query\_valid} mask; it does not consume the first prediction, the Future logits, or the global slots. The global-slot states are internal decoder memory and are not part of the Transition input.

\subsection{Topology-Constrained Transition Network}
\label{app:transition}

The Transition Network first reads the current context without seeing the candidate set. Its input is the history, the destination token, and the downstream part of the Future representation:
\begin{figure}[t]
    \centering
    \includegraphics[width=\linewidth]{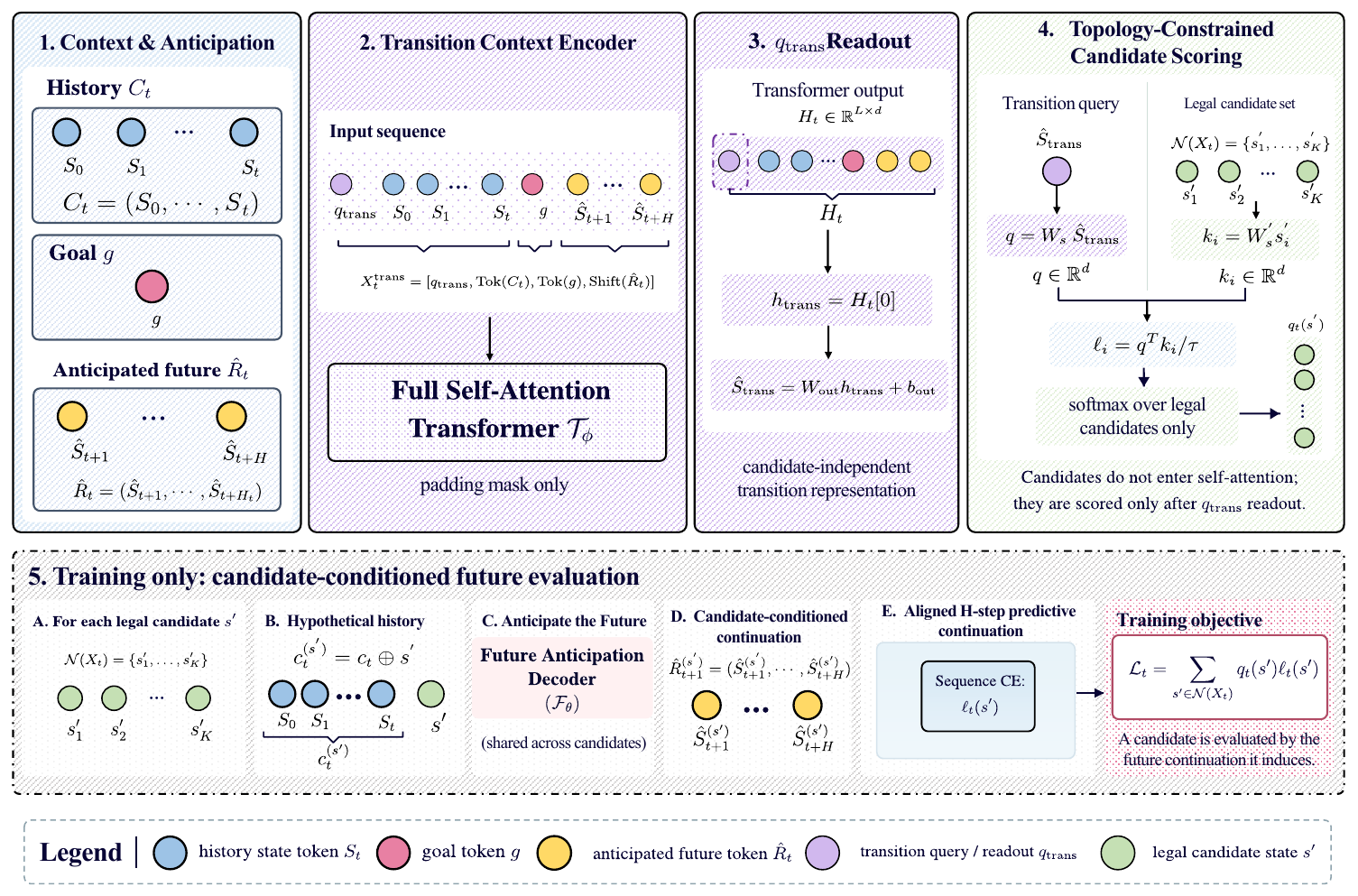}
    \caption{Topology-constrained Transition Network. A candidate-independent readout summarizes $c_t$, $g$, and the downstream Future hidden tail. Complete legal node--fiber candidates are scored afterward with a projected dot product and normalized only over $\mathcal U_t$; the STOP event uses its independent scalar head. Candidate-conditioned future evaluation is a separate branch performed after the base distribution.}
    \label{fig:app-transition-network}
\end{figure}

\begin{equation}
    X_t^{\mathrm{tr}}
    = [q_{\mathrm{trans}},\,\operatorname{Proj}_C(c_t),\,
       \operatorname{Proj}_g(g),\,\operatorname{Proj}_F(Z_t^F[2{:}H])].
    \label{eq:app-transition-input}
\end{equation}
Here $q_{\mathrm{trans}}$ is a learned readout token. The Transition Network applies full self-attention to this sequence and uses padding masks for invalid history or future positions. Candidate states are not appended to this sequence, so changing the number or order of legal candidates cannot change the contextual readout. With
\begin{equation}
    H_t^{\mathrm{tr}}=\operatorname{Transformer}_{\phi}
    (X_t^{\mathrm{tr}};\operatorname{mask}_t),
    \qquad
    h_t^{\mathrm{tr}}=H_t^{\mathrm{tr}}[q_{\mathrm{trans}}],
    \label{eq:app-transition-readout}
\end{equation}
the network produces one context-dependent prediction in the complete state space:
\begin{equation}
    \begin{aligned}
    r_t^{\mathrm{tr}} &= W_{\mathrm{out}}h_t^{\mathrm{tr}}+b_{\mathrm{out}},
    & q_t^{\mathrm{key}} &= W_qr_t^{\mathrm{tr}},\\
    h_{t+1}^{(u)} &= W_{\mathrm{cand}}[z_u\Vert A_{t+1}^{(u)}]+e_{\mathrm{cand}},
    & k_t^{(u)} &= W_kh_{t+1}^{(u)}.
    \end{aligned}
    \label{eq:app-transition-projections}
\end{equation}
The representation $r_t^{\mathrm{tr}}$ is a prediction in the complete node--fiber state space. The candidate encoder uses the same node and fiber features as the online state construction, so candidate $u$ is represented by its node embedding and incoming fiber $A_{t+1}^{(u)}$. The notation $q_t^{\mathrm{key}}$ denotes a projected query vector and is distinct from the event distribution $q_t$. For each legal movement candidate $s_{t+1}^{(u)}$, the projected dot-product score is
\begin{equation}
    \ell_t\bigl(s_{t+1}^{(u)}\bigr)
    =\frac{(q_t^{\mathrm{key}})^\top k_t^{(u)}}{\kappa},
    \qquad \kappa>0.
    \label{eq:app-transition-movement-score}
\end{equation}
STOP has no node or fiber state. When it is admissible, its score is produced by an independent learned scalar head,
\begin{equation}
    \ell_t(\mathrm{STOP})
    =w_{\mathrm{STOP}}^\top h_t^{\mathrm{tr}}+b_{\mathrm{STOP}}.
    \label{eq:app-transition-stop-score}
\end{equation}
The legal event support is
\begin{equation}
 \mathcal U_t=
 \begin{cases}
  \mathcal U_t^{\mathrm{move}}, & X_t\ne g,\\[2pt]
  \mathcal U_t^{\mathrm{move}}\cup\{\mathrm{STOP}\}, & X_t=g,
 \end{cases}
  \qquad
  \mathcal U_t^{\mathrm{move}}
  =\{s_{t+1}^{(u)}:u\in\mathcal N_{\mathcal G}^+(X_t)\}.
  \label{eq:app-legal-support}
\end{equation}
The event distribution is the masked softmax over exactly this set:
\begin{equation}
    q_t(e)
    =\frac{\exp\{\ell_t(e)\}}
    {\displaystyle\sum_{v\in\mathcal U_t}\exp\{\ell_t(v)\}},
    \qquad e\in\mathcal U_t,
    \label{eq:app-transition-distribution}
\end{equation}
and $q_t(e)=0$ outside the support. Padding entries and illegal candidates therefore never contribute to the denominator. In particular, the STOP column is masked to $-\infty$ whenever $X_t\ne g$; it becomes a competing event at the goal, and is the unique event when the goal has no outgoing movement. A non-goal state with no legal outgoing movement has an empty support and is reported as an invalid transition rather than being forced to stop.

The number of legal candidates can vary across the road network. For batching, candidate rows may be padded to a common width, but padding is masked before scoring and never enters the normalizer. The support remains exactly $\mathcal U_t^{\mathrm{move}}$ (together with STOP when it is admissible); no sampled or synthetic candidate changes the legal event set.

The output interface is consequently
\begin{equation}
    (c_t,g,Z_t^F[2{:}H],\mathcal U_t^{\mathrm{move}})
    \longmapsto
    \bigl(r_t^{\mathrm{tr}},\{\ell_t(e)\}_{e\in\mathcal U_t},q_t\bigr).
    \label{eq:app-transition-interface}
\end{equation}
Candidate-conditioned branches are evaluated only after this base readout and distribution have been formed. They can refine inference when enabled, but they do not change the candidate-independent definition or legal support of $q_t$.

\subsection{Candidate-Conditioned Re-Anticipation}
\label{app:reanticipation}
\label{app:execution}

The base distribution $q_t$ scores legal events from the current history and its Future representation. An optional inference refinement asks whether each proposed movement remains compatible with the model's own forecast after that movement is appended. For a legal movement $s'\in\mathcal U_t^{\mathrm{move}}$, form a hypothetical branch by appending its full node--fiber state to a copy of $c_t$:
\begin{equation}
 c_t^{(s')}=c_t\mathbin{\Vert}s',\qquad
 R_{t+1}^{(s')}=\mathcal F_{\theta}(c_t^{(s')},g)
 =\bigl(Z_{t+1}^{F,(s')},L_{t+1}^{(s')}\bigr).
 \label{eq:app-candidate-branch}
\end{equation}
The branch uses the same Future Anticipation Decoder parameters and the absolute query positions associated with the extended history. It is hypothetical: evaluating it does not alter $c_t$. Candidate branches can be flattened over the active history--candidate pairs and decoded in a single batched call; padded or illegal candidates are omitted.

Appending one state shifts the forecast window by one position. Thus the current forecast at query index $h$ and the branch forecast at index $h-1$ refer to the same absolute future position for $h=2,\ldots,H$. Figure~\ref{fig:app-branch-alignment} shows this overlap; the first query in each row is excluded from the comparison.
\begin{figure}[t]
\centering
\small
\setlength{\tabcolsep}{5pt}
\begin{tabular}{c|c|ccccc}
 & context & query 1 & query 2 & query 3 & $\cdots$ & query $H$ \\
\hline
 Current & $c_t$ & $t+1$ & $t+2$ & $t+3$ & $\cdots$ & $t+H$ \\
 Branch & $c_t\Vert s'$ & $t+2$ & $t+3$ & $t+4$ & $\cdots$ & $t+H+1$ \\
\end{tabular}
\[
\underbrace{L_t[2],\ldots,L_t[H]}_{\text{current: }t+2,\ldots,t+H}
\quad\longleftrightarrow\quad
\underbrace{L_{t+1}^{(s')}[1],\ldots,L_{t+1}^{(s')}[H-1]}_{\text{branch: }t+2,\ldots,t+H}.
\]
\caption{Overlap alignment for candidate-conditioned re-anticipation. Predictions are compared only when they refer to the same absolute future position.}
\label{fig:app-branch-alignment}
\end{figure}

Let $p_t[h]=\operatorname{softmax}(L_t[h])$ and $p_{t+1}^{(s')}[h]=\operatorname{softmax}(L_{t+1}^{(s')}[h])$, over the shared future event alphabet $\mathcal Y$ (node events and STOP). For $H>1$, the optional consistency energy is the mean Jensen--Shannon divergence over the $H-1$ overlapping positions:
\begin{equation}
 E_{\mathrm{fut}}(s')=
 \frac{1}{H-1}\sum_{h=2}^{H}
 \operatorname{JSD}\!\left(p_t[h],p_{t+1}^{(s')}[h-1]\right).
 \label{eq:app-future-consistency}
\end{equation}
Here $\operatorname{JSD}(p,r)=\tfrac12\operatorname{KL}(p\Vert m)+\tfrac12\operatorname{KL}(r\Vert m)$, where $m=\tfrac12(p+r)$.
This score uses predicted distributions only; it does not use realized future labels or \texttt{target\_valid}. If $H\leq1$, no positions overlap and the future energy is defined as zero, so no empty average is formed. STOP has no successor state and is assigned zero future energy.

When the inference weight $w>0$ (the default configuration uses $w=1$), combine this optional score with the base event probability:
\begin{equation}
 E_t(e)=-\log q_t(e)+wE_{\mathrm{fut}}(e),\qquad
 \pi_t(e)=\frac{\exp[-E_t(e)]}
 {\sum_{v\in\mathcal U_t}\exp[-E_t(v)]},\quad e\in\mathcal U_t.
 \label{eq:app-reranked-policy}
\end{equation}
Here $E_{\mathrm{fut}}(e)$ is the branch score for a movement and zero for STOP. Greedy decoding selects the legal event with minimum $E_t$; sampling draws from $\pi_t$. Setting $w=0$ (or $H\leq1$) uses the base distribution $q_t$ directly. After selection, only the chosen movement state is appended to the realized history; the next decision computes a fresh Future window from that updated history.

\subsection{Closed-loop inference and termination}
\label{app:closed-loop}

Foresight generates a trajectory by repeatedly anticipating, selecting one graph-valid event, executing it, and revising the anticipation from the resulting history. The future window is reconstructed at every decision. In particular, the previous prediction is not shifted forward after a movement: the new history determines new query positions, masks, candidate fibers, and decoder states. The procedure below uses $K_{\max}$ as an external decision budget; the STOP event counts as one decision.

\begin{figure}[t]
    \centering
    \includegraphics[width=\linewidth]{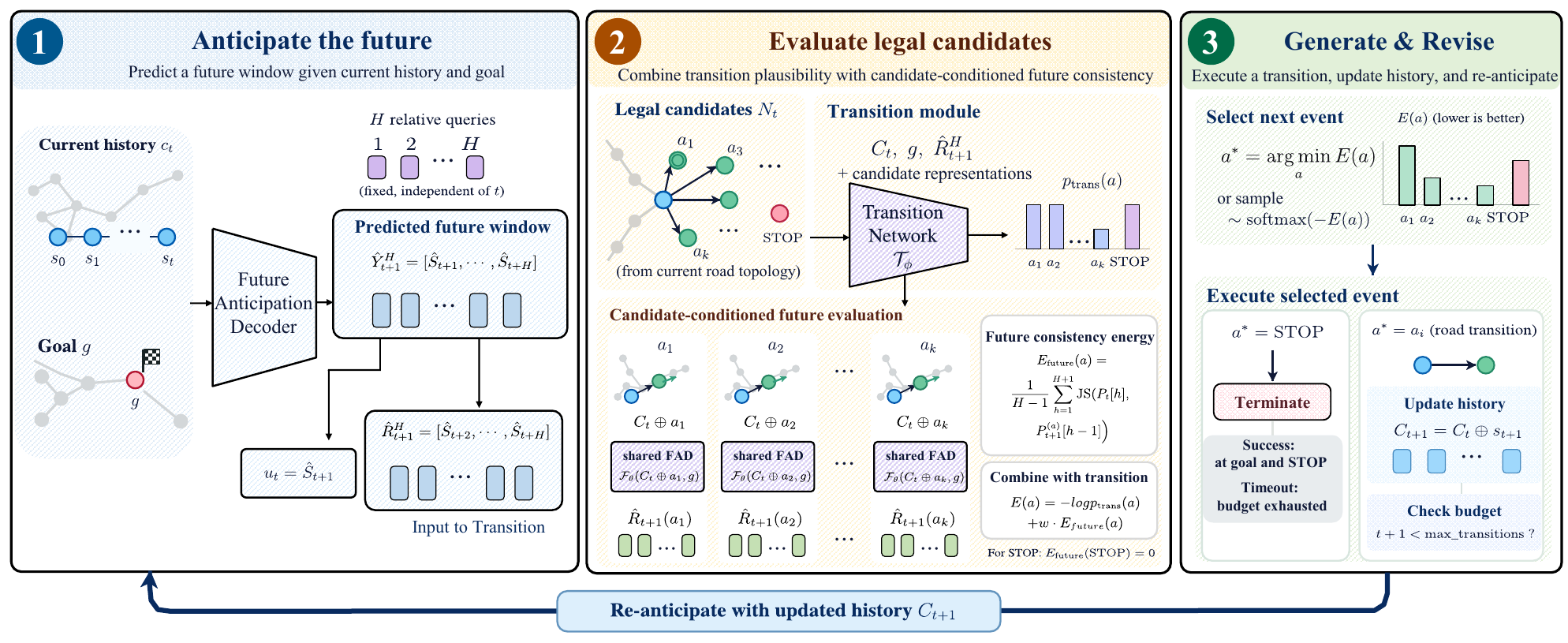}
    \caption{Closed-loop generation. At each decision, Foresight anticipates a fixed future window from the realized history and goal, scores the graph-legal events, optionally evaluates candidate-conditioned future consistency, and selects one movement or STOP. A movement updates the history and triggers a fresh Future Anticipation Decoder call; STOP terminates only at the goal, while budget exhaustion is reported as a timeout.}
    \label{fig:app-generation}
\end{figure}

The data flow at one decision is therefore
\[
 (c_k,g)\;\longrightarrow\;\mathcal F_{\theta}
 \;\longrightarrow\;q_k
 \;\longrightarrow\;\pi_k\;(\text{optional})
 \;\longrightarrow\;\text{one movement or STOP}.
\]
The current Future forward and all candidate branches use shared decoder parameters. Branches are evaluated only for valid movement candidates; padding rows are removed before decoding and never enter a probability denominator. The branch batch may be flattened over active request--candidate pairs, but the real trajectory remains sequential because $c_{k+1}$ depends on the event selected at step $k$.

STOP and failure states have distinct meanings. STOP is admissible only at the destination, does not create a successor state, and terminates the rollout only when it is selected. Visiting the destination does not terminate generation; a trajectory may leave the destination if a movement is selected there. Consequently, an initial history whose current node is already $g$ still invokes the model and must select STOP. A non-destination node with no legal outgoing movement has an empty event support and returns \texttt{invalid\_transition}; STOP is never used as a fallback. If the budget is exhausted before STOP is selected, the result is \texttt{timeout} (or truncation under the reporting terminology). Non-finite model outputs or an invalid sampled event are reported as \texttt{numerical\_error} rather than silently repaired.

\begin{table}[t]
\centering
\small
\caption{Termination and boundary semantics during generation.}
\label{tab:app-termination-semantics}
\begin{tabular}{p{0.38\linewidth}|p{0.49\linewidth}}
\toprule
Current condition & Event support and outcome \\
\midrule
Non-goal with legal outgoing edges & Movement candidates only; STOP is masked. \\
Goal with legal outgoing edges & Movement candidates and STOP compete under $q_t$ (or $\pi_t$). \\
Goal with no outgoing edges & STOP is the unique legal event. \\
Non-goal with no outgoing edges & Empty support; return \texttt{invalid\_transition}. \\
STOP selected at the goal & End without appending a state; \texttt{success=true}. \\
Budget exhausted before STOP & End with \texttt{timeout}; \texttt{success=false}. \\
\bottomrule
\end{tabular}
\end{table}

The diagnostic flag \texttt{goal\_arrived} records whether the realized history has ever visited $g$; it is independent of the terminal event. A rollout is successful only when STOP is selected while the current node is $g$, so merely visiting $g$ does not imply success. This matches the stopping-time convention in the theory appendix, where termination is induced by the selected STOP event rather than by the first visit to the destination.

For a batch of requests, generation proceeds by an active-wavefront schedule. At each depth, only requests that have not returned \texttt{stop}, \texttt{invalid\_transition}, \texttt{timeout}, or \texttt{numerical\_error} are padded and passed through the shared Future and Transition forwards. Candidate branches are flattened across the active requests and legal movement candidates, then scattered back to their request rows before selection. Each request keeps its own history, seed, and sampling state; removing a finished request from later waves does not alter the ordering of the remaining candidates. A trace records the node sequence and complete states, legal candidates, $q_t$ (and $\pi_t$ when used), selected event, query/mask metadata, consistency energy, elapsed time, and the final stop reason.

\subsection{Architecture Configuration and Inference Complexity}
\label{app:model-configurations}

The scale names $s1$--$s5$ refer to the resolved architecture presets in
\texttt{configs/rest\_s1}--\texttt{configs/rest\_s5}. The dimensions and
depths below are transcribed from the current scale specification; they
describe the network instance and do not determine the optimizer, training
schedule, or generation budget. The Future Anticipation Decoder uses width
$d_F=d_{\mathrm{model}}=d_X$, while the Transition Network uses width
$d_T=d_X$. All production scales use $d_V=64$, fiber hidden width 128,
sinusoidal time encodings, zero dropout, eight Future attention heads, four
Transition heads, two-layer base/fiber MLPs with base hidden width $2d_X$,
and Transformer FFN widths four times their model width. Every scale uses the
fixed Future horizon $H=64$; the realized suffix length is represented by
the validity mask.

\begin{table}[t]
\centering
\small
\caption{Resolved architecture dimensions for the five model scales. $H$ is fixed at 64 in every configuration.}
\label{tab:app-architecture-config}
\setlength{\tabcolsep}{4pt}
\begin{tabular}{lrrrrrrrrrr}
\toprule
Scale & $d_X=d_F=d_T$ & $d_V$ & base hidden & $M$ & $H$ & $L_E$ & $L_D$ & $L_T$ & FFN ratio\\
\midrule
s1 & 72  & 64 & 144 & 16 & 64 & 2 & 2 & 2 & $4\times$\\
s2 & 96  & 64 & 192 & 24 & 64 & 3 & 2 & 2 & $4\times$\\
s3 & 152 & 64 & 304 & 32 & 64 & 4 & 3 & 3 & $4\times$\\
s4 & 240 & 64 & 480 & 48 & 64 & 6 & 4 & 4 & $4\times$\\
s5 & 400 & 64 & 800 & 64 & 64 & 8 & 6 & 6 & $4\times$\\
\bottomrule
\end{tabular}
\end{table}

The current inference protocol uses the projected dot-product candidate
score with a positive finite temperature $\kappa$. Candidate-conditioned
future consistency is enabled with weight $w=1$ in the default protocol;
setting $w=0$ gives the Transition-only ablation. The selection mode,
sampling temperature, and decision budget are run-level settings and must be
read from the saved \texttt{resolved\_config}; no fixed value is asserted
here. Likewise, the global vocabulary size $|\mathcal Y|$ is supplied by the
city-specific ordered node table and is not determined by the scale name.
The resulting parameter counts therefore vary by city. The current report is:

\begin{table}[t]
\centering
\small
\caption{Total trainable parameters (Future plus Transition) by city and model scale. Counts include the Future vocabulary head and exclude optimizer state, gradients, activations, and runtime caches.}
\label{tab:app-parameter-counts}
\setlength{\tabcolsep}{3pt}
\begin{tabular}{lrrrrr}
\toprule
City ($|\mathcal V|$) & s1 & s2 & s3 & s4 & s5\\
\midrule
Beijing (40,060) & 3,300,102 & 4,628,670 & 8,448,638 & 17,366,670 & 44,881,710\\
Porto (11,024) & 1,180,474 & 1,812,178 & 4,006,130 & 10,368,994 & 33,238,274\\
San Francisco (27,187) & 2,360,373 & 3,379,989 & 6,479,069 & 14,264,277 & 39,719,637\\
\bottomrule
\end{tabular}
\end{table}

Let $N_t$ be the number of valid history tokens, $D_t$ the number of legal
movement candidates, and $V=|\mathcal Y|$ the global predictive vocabulary.
One decision performs one current Future forward and one Transition forward.
With dense attention, the current Future cost is
$\mathcal O((N_t+H+M+2)^2d_F+HVd_F)$, while the Transition cost is
$\mathcal O((N_t+H-1)^2d_T+D_td_T)$ up to layer and head factors. When
future-consistency reranking is enabled, the valid movement branches add a
batched Future computation of
$\mathcal O(D_t(N_t+H+M+2)^2d_F+D_tHVd_F)$; batching reduces kernel-launch
overhead but does not remove this candidate-dependent work or its activation
memory. Candidate branches are flattened over active request--candidate
pairs, whereas different realized steps remain sequential because the next
history depends on the selected event.

Static node encodings may be cached when model parameters are fixed. Fiber
features remain online functions of each source--destination pair, and all
history-conditioned Future hidden states and logits are recomputed after a
real movement. The implementation therefore does not claim a cache for the
entire decoder state or a fixed latency improvement from batching. Measured
latency, throughput, and peak memory are run-dependent quantities and belong
to the experimental protocol (Appendix~\ref{app:experimental-setup-protocol}). Architecture
and parameter-scale specifications are reported in Appendix~\ref{app:model-configuration};
the result tables in Appendix~F should be read together with
Tables~\ref{tab:app-architecture-config} and~\ref{tab:app-parameter-counts}.

\section{Experimental Setup and Evaluation Protocol}
\label{app:experiments}

This appendix specifies the data, model configuration, training procedure, and
evaluation protocol used for the reported experiments. Porto, Beijing, and San
Francisco use the official HOSER road graphs and train/validation/test splits
\citep{cao2025holistic}. Training reads only the training split; validation is
used for checkpoint selection and development-time model selection; the test
split is read only after the checkpoint and test-condition manifest have been
fixed. All methods use the same graph, test conditions, replicate policy,
generation budget, and failure-accounting rule. The fixed settings used for
the reported runs are stated below.
Complete numerical evidence is reported in Appendix~\ref{app:full-results},
while generated spatial and trajectory examples are reported separately in
Appendix~\ref{app:full-visuals}.

\subsection{Datasets and Baselines}
\label{app:datasets-baselines}

\paragraph{Dataset Description and Sources}
\label{app:dataset-description}

We evaluate goal-conditioned road-network trajectory generation on the Porto,
Beijing, and San Francisco subsets of the public HOSER dataset. The supplied
road graph provides directed road states, road attributes, and legal outgoing
transitions; trajectory files provide the observed road-state sequences. The
three subsets represent the corresponding urban road regions, and the city
name is retained as the geographic unit throughout training and evaluation.
The road graph and trajectory identifiers are retained when constructing the
experimental records, so that a generated path can be checked against the
same graph and reference representation used by the test evaluator.

Preprocessing performs the dataset-provided road-to-trajectory association and
removes records that cannot be represented as graph-valid directed sequences
or that contain non-finite required attributes. Continuous-feature
normalization and any transition statistics are fitted on the training split
and then reused without refitting on validation or test. We do not introduce a
new random split or pool trajectories across cities. Dataset-specific
engineering details are kept in the data bundle; this appendix reports only
the choices that affect the scientific comparison.

\paragraph{Dataset Splits and Statistics}
\label{app:dataset-splits}

The official split is used independently for each city. Training trajectories
are used to fit model parameters and training-only statistics. Validation
trajectories are fixed while selecting a checkpoint and comparing development
settings. Test trajectories and their reference paths are held out until the
checkpoint, condition manifest, generation seeds, and evaluation configuration
are frozen. A test condition contains the origin/goal request and its linked
reference metadata; several conditions may share one OD pair.

Table~\ref{tab:dataset-statistics} is the compact dataset summary, following
the statistics reported for the public HOSER data in Cao et al.~\citep{cao2025holistic}.
Road counts, trajectory totals, average distances, and average durations are
copied from that dataset report. Split counts are properties of the accepted
experimental manifest and are reported separately when available.

\begin{table}[t]
\centering
\caption{Dataset statistics for the three evaluation cities, following the
public HOSER dataset report~\citep{cao2025holistic}.}
\label{tab:dataset-statistics}
\setlength{\tabcolsep}{9pt}
\begin{tabular}{lrrrrr}
\toprule
Dataset & Roads & Trajectories & Avg. distance & Avg. time \\
\midrule
Porto & 11,024 & 687,656 & 3.66\,km & 7.99\,min \\
Beijing & 40,060 & 899,115 & 5.16\,km & 12.87\,min \\
San Francisco & 27,187 & 293,023 & 3.41\,km & 9.29\,min \\
\bottomrule
\end{tabular}
\end{table}

Every method is evaluated on the same ordered condition manifest. The manifest
fixes the origin, destination, any additional goal-conditioned fields, the
reference identifier, and the requested replicate seeds. No test trajectory
is used to fit normalization, transition frequencies, or a checkpoint.

\paragraph{Baseline Models}
\label{app:baseline-models}

Baselines cover three comparison families. \emph{Statistical references}
include Markov transition sampling, which uses only the training split. \emph{Graph-search references} include Dijkstra~\citep{dijkstra1959note} and
Yen's $k$-shortest-path procedure~\citep{Yen1971find}, run on the same directed
graph and OD requests. \emph{Learned generators} include Semantic
Routing~\citep{zhaosemantic}, HOSER~\citep{cao2025holistic}, TS-TrajGen~\citep{ContinuousTrajectoryGene},
GDP~\citep{shi2024graphconstrained}, Diff-RNTraj~\citep{Wei2024diffrntraj},
DiffTraj~\citep{zhu2023difftraj}, TrajFlow~\citep{li2026trajflow}, and
Seed~\citep{rao2025seed}, matching the methods listed in the main experiment
table.

For each learned baseline, the original representation and decoder are kept
as the method's defining component. If the original method is not
OD-conditioned, we use an explicitly documented OD adaptation: the origin and
destination are supplied through its conditioning interface (or its input
representation), and the adaptation is trained on the same city and split.
Continuous GPS outputs are mapped to the common road representation with the
fixed map-matching rule; conversion failures remain failed attempts. Search
baselines use fixed edge costs, $k$, candidate ordering, and search budgets.
All baselines use the same test conditions, replicate seeds, maximum
generation budget, and failure denominator. Thus differences due to the
baseline's original modeling idea are separated from choices made solely to
make the comparison executable under the common protocol.

More specifically, Semantic Routing~\citep{zhaosemantic} and HOSER~\citep{cao2025holistic} are stepwise road-level
navigational generators whose road-token or semantic representation is kept
in its native form; their goal interface is used directly when available. GDP
~\citep{shi2024graphconstrained} and Diff-RNTraj~\citep{Wei2024diffrntraj} generate graph-constrained paths or road-network trajectories
with diffusion, so their graph support is retained and the OD request is
added as the declared conditioning input when required. TS-TrajGen uses a
two-stage continuous trajectory generator~\citep{ContinuousTrajectoryGene}, while DiffTraj~\citep{zhu2023difftraj}, TrajFlow~\citep{li2026trajflow}, and Seed~\citep{rao2025seed}
operate on continuous GPS, flow, or sequence--diffusion representations. For
these methods, the only comparison adaptation is OD conditioning plus the
fixed GPS-to-road conversion; no generated path is repaired by a shortest-path
replacement. Markov estimates local transition frequencies from training
trajectories, and the search references decode their fixed candidate sets deterministically before
the common replicate and failure accounting is applied.

\subsection{Experimental Setup and Evaluation Protocol}
\label{app:experimental-setup-protocol}

\paragraph{Model Configuration and Parameter Scale}
\label{app:model-configuration}

Foresight uses the shared node--fiber state representation, a structured
Future Anticipation Decoder, and a topology-constrained Transition Network
(Appendix~\ref{app:architecture}). The scale name identifies one resolved
configuration directory, while the city-specific graph bundle supplies the
global road vocabulary. The complete per-scale configuration, including all
widths, depths, heads, slot counts, feed-forward dimensions, and the fixed
Future horizon, is given in Table~\ref{tab:model-scale-configuration}. The
Transition readout uses the projected dot-product candidate score described in
Appendix~\ref{app:transition}.

The parameter report counts every trainable tensor by \texttt{numel}. For a
city with vocabulary size $V$, the Future vocabulary head contributes
$(d_X+1)V$ parameters; all other Future parameters and all Transition
parameters are independent of the city. Tables~\ref{tab:model-parameter-statistics}
and~\ref{tab:model-parameter-by-city} give the decomposition and the resulting
city/scale totals. Counts exclude optimizer state, gradients, activations, and
runtime caches. The selected scale, vocabulary size, and full parameter report
must be read from the saved \texttt{resolved\_config} when reproducing a run.

\begin{table}[t]
\centering\normalsize
\caption{Parameter decomposition for the five resolved model scales. The Future vocabulary head is $(d_X+1)V$ for vocabulary size $V$.}
\label{tab:model-parameter-statistics}
\setlength{\tabcolsep}{3pt}
\begin{tabular}{lrrrrr}
\toprule
Scale & Future fixed & Future head & Transition & Total formula \\
\midrule
s1 & 199,401 & $73V$ & 176,321 & $375,722+73V$ \\
s2 & 438,657 & $97V$ & 304,193 & $742,850+97V$ \\
s3 & 1,303,273 & $153V$ & 1,016,185 & $2,319,458+153V$ \\
s4 & 4,526,385 & $241V$ & 3,185,825 & $7,712,210+241V$ \\
s5 & 16,201,585 & $401V$ & 12,616,065 & $28,817,650+401V$ \\
\bottomrule
\end{tabular}
\end{table}

\begin{table}[t]
\centering
\normalsize
\caption{Total trainable parameters by city and model scale, reported in millions (M). $V$ is the graph vocabulary size; counts include the Future vocabulary head.}
\label{tab:model-parameter-by-city}
\setlength{\tabcolsep}{3pt}
\begin{tabular}{lrrrrr}
\toprule
City ($V$) & s1 & s2 & s3 & s4 & s5 \\
\midrule
Beijing (40,060) & 3.300 & 4.629 & 8.449 & 17.367 & 44.882 \\
Porto (11,024) & 1.180 & 1.812 & 4.006 & 10.369 & 33.238 \\
San Francisco (27,187) & 2.360 & 3.380 & 6.479 & 14.264 & 39.720 \\
\bottomrule
\end{tabular}
\end{table}

\begin{table}[t]
\centering
\normalsize
\caption{Complete resolved configuration for each model scale. The columns are split into two panels for readability; parameters not varying with scale are repeated explicitly.}
\label{tab:model-scale-configuration}
\setlength{\tabcolsep}{4pt}
\textbf{Panel A: widths, slots, and horizon}\\[2pt]
\begin{tabular}{lrrrrrr}
\toprule
Scale & $d_X=d_{\rm model}$ & $d_V$ & base hid. & fiber hid. & slots & $H$ \\
\midrule
s1 & 72  & 64 & 144 & 128 & 16 & 64 \\
s2 & 96  & 64 & 192 & 128 & 24 & 64 \\
s3 & 152 & 64 & 304 & 128 & 32 & 64 \\
s4 & 240 & 64 & 480 & 128 & 48 & 64 \\
s5 & 400 & 64 & 800 & 128 & 64 & 64 \\
\bottomrule
\end{tabular}
\\[8pt]
\textbf{Panel B: Base/Fiber and Future Decoder}\\[2pt]
\begin{tabular}{lrrrrr}
\toprule
Scale & Base layers & Fiber layers & Future heads & Future layers & Future FFN \\
\midrule
s1 & 2 & 2 & 8 & 2 & 288 \\
s2 & 2 & 2 & 8 & 3 & 384 \\
s3 & 2 & 2 & 8 & 4 & 608 \\
s4 & 2 & 2 & 8 & 6 & 960 \\
s5 & 2 & 2 & 8 & 8 & 1,600 \\
\bottomrule
\end{tabular}
\\[8pt]
\textbf{Panel C: Transition Network}\\[2pt]
\begin{tabular}{lrrr}
\toprule
Scale & Transition heads & Transition layers & Transition FFN \\
\midrule
s1 & 4 & 2 & 288 \\
s2 & 4 & 2 & 384 \\
s3 & 4 & 3 & 608 \\
s4 & 4 & 4 & 960 \\
s5 & 4 & 6 & 1,600 \\
\bottomrule
\end{tabular}
\end{table}

\paragraph{Training Configuration}
\label{app:training-configuration}

Training minimizes the unified candidate-conditioned sequence objective. Future
targets are stop-gradient supervision targets; they are never exposed as
inputs to the Future Decoder or Transition Network. The best checkpoint is
selected by the fixed validation loss in evaluation mode and then frozen
before test generation. The values below are the resolved run configuration
used for the reported experiments.

\begin{table}[t]
\centering\small
\caption{Training configuration used for the reported experiments.}
\label{tab:training-configuration}
\begin{tabular}{p{.46\linewidth}p{.40\linewidth}}
\toprule
Setting & Value \\
\midrule
Optimizer & AdamW \\
Learning rate & 5e-4 \\
Weight decay & 1e-4 \\
Learning-rate schedule & constant \\
Warmup steps & 3000 \\
Total optimizer updates & 60000 \\
Trajectories (paths) per trajectory batch & 64 \\
Anchor micro-batch size & 512 \\
Outer gradient accumulation & 1 \\
Anticipation horizon $H$ & 64 \\
Gradient clipping and numerical precision & float 32 \\
Checkpoint selection & minimum fixed validation loss \\
\bottomrule
\end{tabular}
\end{table}

The \emph{trajectory batch} is the complete set of sampled trajectories from
which valid decision anchors are constructed. The \emph{anchor micro-batch}
is only a memory-scheduling chunk inside that trajectory batch; it does not
define a different data distribution. Outer gradient accumulation, if used,
is a separate update boundary across trajectory batches. All global
normalizers are computed before chunking (with separate Future and Transition
valid-anchor counts when the objective uses separate denominators), and each
chunk contributes its share to the same batch gradient. Consequently, anchor
micro-batching changes memory use and graph lifetime but does not change the
mathematical loss or its normalization.

\paragraph{Evaluation Metrics}
\label{app:evaluation-metrics}

All metrics operate on the ordered 2-D road trajectory representation obtained
from the HOSER road graph (road-state geometry mapped to a local metric
coordinate system). Spatial distances are measured in kilometres; generated
and reference trajectories are compared only when both have valid
representations. We report two complementary metric families. \emph{Distribution-level
metrics} are Distance JSD and Radius JSD. Distance JSD is the Jensen--Shannon
divergence between the reference and generated distributions of trajectory
travel distance, while Radius JSD compares their distributions of spatial
radius from the trajectory origin. Both use the same fixed city-specific
support and bins for every method, are reported in the tables in units of
$10^{-2}$, and are lower-is-better.

\emph{Trajectory-level similarity metrics} compare each generated trajectory
with its matched reference trajectory. Hausdorff distance~\citep{huttenlocher1993comparing}
measures the maximum spatial deviation between the two point sets; DTW~\citep{sakoechiba1978dynamic}
uses the standard dynamic-programming alignment to accumulate the minimum
sequence-level Euclidean distance; and EDR~\citep{chen2005robust} is an
edit-distance-style similarity that matches points within a fixed tolerance
$\epsilon=0.05$\,km (50\,m). Hausdorff and DTW are
reported as accumulated distances in kilometres, whereas EDR is divided by
the reference sequence length and therefore lies on a normalized scale.
All three are lower-is-better. The implementation uses the standard
definitions above, with only the stated road-trajectory conversion,
city-level binning, and normalization applied.

\paragraph{Test and Generation Procedure}
\label{app:test-generation}

After checkpoint selection, generation runs in evaluation/no-gradient mode on
the frozen checkpoint and the fixed test-condition manifest. A condition
provides the origin, destination, and any declared goal-conditioning fields;
the generator reads the generated prefix, graph, legal candidates, and its own
predictions, never a true future suffix, reference label, or true path length.
For each condition, the manifest requests a fixed number $K_{\mathrm{rep}}$ of
replicates with persisted generation seeds. The reported runs use 5,000 fixed
test conditions and generation seeds
$\{0,1,2,3,4\}$. Decoding follows the declared run-level greedy or sampling configuration shared
by all methods where applicable.

At each transition, the model scores graph-legal movement events and the
independent \texttt{STOP} event. \texttt{STOP} is admissible only when the
current node is the destination. The maximum transition budget
$K_{\max}$ is fixed at 256 transitions. A rollout ends at explicit
\texttt{STOP}, at
$K_{\max}$, or at an abnormal failure such as an empty non-goal support or a
non-finite model output. Reaching the destination alone does not terminate a
rollout. Every attempt, including timeout and failure records, remains in the
evaluation corpus with its condition ID, OD ID, replicate seed, node
sequence, stop reason, success flag, and error code. Completion rates use all attempts as the denominator.

\paragraph{OD Bootstrap Evaluation}
\label{app:od-bootstrap}

The primary uncertainty protocol treats an exact \texttt{od\_id} as one
cluster. Generated replicates and conditions are first grouped by that OD;
when an OD contains multiple conditions, the condition-level summaries are
averaged within the OD before resampling. A bootstrap replicate samples OD
clusters with replacement, recomputes every reported metric from the sampled
clusters, and stores the resulting statistic. The reported point estimate is
the statistic on the complete corpus, and the uncertainty interval or
bootstrap standard deviation is obtained from the bootstrap distribution. The
reported results use 1,000 resamples with bootstrap seed 0.

Resampling OD clusters keeps repeated conditions from the same origin--goal
pair from being treated as independent evidence. Distance JSD, Radius JSD,
Hausdorff, EDR, DTW, and completion diagnostics are recomputed under the same
fixed supports, matching rules, and denominator definitions in every
bootstrap replicate. Undefined metrics remain unavailable and are never
zero-filled.

\paragraph{Condition Bootstrap Evaluation}
\label{app:condition-bootstrap}

The condition-bootstrap protocol uses the exact evaluation condition as its
resampling unit. A condition is the frozen origin/goal request together with
its reference metadata and all requested generated replicates. Conditions
that share an OD pair remain distinct clusters here, because the purpose is to
measure sensitivity to condition weighting. A bootstrap replicate samples
condition IDs with replacement and retains all generated replicates belonging
to each sampled condition as one unit before recomputing the metrics.

Thus, OD bootstrap answers how results vary across origin--goal pairs, whereas
condition bootstrap gives every supplied goal-conditioned test condition its
own sampling weight. Both protocols preserve the condition-to-replicate
grouping and the complete failure corpus. HOSER-style file-level mean
$\pm$ standard deviation is retained only as a supplementary compatibility
statistic: it computes one summary per generation file and reports variation
across files, without replacing the OD-bootstrap protocol used for the main
evidence.

\subsection{PyTorch-style Pseudocode}
\label{app:pseudocode}

The four algorithms below state the tensor and control flow of Foresight at
the architecture level. They omit repository-specific APIs and file paths.
The Future Decoder predicts over the global alphabet
$\mathcal Y=\mathcal V\cup\{\mathrm{STOP}\}$, whereas the Transition Network
normalizes only over the current graph-legal support. A future target mask
affects supervision only; it is not supplied as a conditioning feature.

\paragraph{Future Anticipation Decoder}
\label{app:pseudocode-future-decoder}

\begin{algorithm}[t]
\caption{Future Anticipation Decoder}
\label{alg:future-decoder}
\begin{algorithmic}[1]
\State \textbf{Input:} realized history $c_t$, destination $g$, query-valid mask $M_t$
\State $C \gets \texttt{encode\_history}(c_t)$; $G \gets \texttt{encode\_goal}(g)$
\State $Q^F \gets \texttt{future\_queries}(H,t)$; $Q^G \gets \texttt{global\_slots}()$
\State $X \gets \texttt{concat}(C,Q^F,G,Q^G,\texttt{STOP\_slot})$
\State $Z \gets \texttt{structured\_transformer}(X,\texttt{visibility\_mask}(c_t,M_t))$
\State $Z^F \gets \texttt{future\_block}(Z)$
\State $L^F \gets \texttt{future\_head}(Z^F) \in \mathbb R^{H\times|\mathcal Y|}$
\State $L^F \gets \texttt{mask\_invalid\_query\_positions}(L^F,M_t)$
\State \Return $Z^F,L^F$
\end{algorithmic}
\end{algorithm}

The structured mask allows each future query to read the valid history,
destination, global slots, and preceding future queries while preventing later
queries or target suffixes from leaking into earlier positions. The first
future hidden state represents the immediate next event; the downstream tail
$Z^F_{2:H}$ is exposed to the Transition Network.

\paragraph{Transition Network}
\label{app:pseudocode-transition-network}

\begin{algorithm}[t]
\caption{Topology-constrained Transition Network}
\label{alg:transition-network}
\begin{algorithmic}[1]
\State \textbf{Input:} history $c_t$, goal $g$, downstream anticipation $Z^F_{2:H}$, graph $G$
\State $\mathcal U_t^{\mathrm{move}} \gets \texttt{legal\_neighbors}(G,c_t[-1])$
\State $u_t \gets \texttt{context\_encoder}(c_t,g,\texttt{detach}(Z^F_{2:H}))$
\State $e_i \gets \texttt{candidate\_encoder}(s_i')$ for each $s_i'\in\mathcal U_t^{\mathrm{move}}$
\State $\ell_i \gets \texttt{projected\_dot}(u_t,e_i)$ for each legal candidate
\State $\ell_{\mathrm{stop}} \gets \texttt{stop\_head}(u_t)$
\State $S_t \gets \mathcal U_t^{\mathrm{move}} \cup \{\mathrm{STOP}\ \text{if }c_t[-1]=g\}$
\State $\ell \gets \texttt{mask\_outside\_support}(\texttt{append}(\ell_i,\ell_{\mathrm{stop}}),S_t)$
\State $q_t \gets \texttt{softmax\_over\_support}(\ell,S_t)$
\State \Return $\ell,q_t,S_t$
\end{algorithmic}
\end{algorithm}

The context readout is formed without appending candidates to the Transformer
sequence, so candidate count and ordering cannot change the context itself.
Only legal movement events and, at the goal, \texttt{STOP} participate in the
normalizer. A non-goal state with no legal movement is reported as an invalid
transition rather than being silently converted to \texttt{STOP}.

\paragraph{Sequential Generation}
\label{app:pseudocode-sequential-generation}

\begin{algorithm}[t]
\caption{Closed-loop Sequential Generation}
\label{alg:sequential-generation}
\begin{algorithmic}[1]
\State \textbf{Input:} condition $(s_0,g)$, graph $G$, transition budget $K_{\max}$
\State $c \gets [s_0]$; $\texttt{goal\_arrived} \gets (s_0=g)$
\For{$t=0,\ldots,K_{\max}-1$}
    \State $(Z^F,L^F) \gets \texttt{FutureDecoder}(c,g,\texttt{all\_valid\_queries})$
    \State $(\ell,q_t,S_t) \gets \texttt{Transition}(c,g,Z^F_{2:H},G)$
    \State $a \gets \texttt{decode}(\ell,q_t,\texttt{configured\_greedy\_or\_sampling})$
    \If{$a=\mathrm{STOP}$}
        \State \textbf{break} \Comment{legal only when the current node is $g$}
    \ElsIf{$a\notin S_t$}
        \State \Return $c,\texttt{invalid\_transition}$
    \Else
        \State $c \gets \texttt{append}(c,\texttt{successor\_state}(a))$
        \State $\texttt{goal\_arrived} \gets \texttt{goal\_arrived}\ \lor\ (c[-1]=g)$
    \EndIf
\EndFor
\If{last action was \texttt{STOP}}
    \State $r \gets \texttt{STOP}$
\Else
    \State $r \gets \texttt{timeout}$
\EndIf
\State \Return $c,r,\texttt{goal\_arrived}$
\end{algorithmic}
\end{algorithm}

The loop makes the receding-anticipation order explicit:
\emph{anticipate $\rightarrow$ select $\rightarrow$ update $\rightarrow$
re-anticipate}. Only the selected movement is realized. Visiting the goal is
recorded separately from selecting \texttt{STOP}; therefore a goal visit does
not by itself mark a successful termination.

\paragraph{Training with Anchor Micro-batching}
\label{app:pseudocode-anchor-training}

For a trajectory batch $B$, let $\mathcal A(B)$ be all valid anchors and let
$Z_B$ denote the global normalizer(s) prescribed by the unified objective (for
example, the total number of valid supervised anchor/position contributions;
Future and Transition terms may have separate counts). These values are
computed before anchors are chunked and are shared by every micro-batch. This
is the key condition that makes micro-batching a memory strategy rather than a
change to the objective.

\begin{algorithm}[t]
\caption{Training with Anchor Micro-batching}
\label{alg:anchor-training}
\begin{algorithmic}[1]
\State \textbf{Input:} trajectory batch $B$, anchor chunk size $m$, optimizer $\mathcal O$
\State $\mathcal A \gets \texttt{build\_valid\_anchors}(B)$; $Z_B \gets \texttt{global\_normalizer}(\mathcal A)$
\State $\mathcal O.\texttt{zero\_grad}()$
\ForAll{$\mathcal A_j \in \texttt{chunk}(\mathcal A,m)$}
    \State $F_j \gets \texttt{FutureDecoder}(\mathcal A_j.\texttt{history},\mathcal A_j.\texttt{goal})$
    \State $L_{F,j} \gets \texttt{masked\_future\_CE}(F_j,\mathcal A_j.\texttt{target})$
    \State $C_j \gets \texttt{legal\_candidates}(\mathcal A_j.\texttt{current\_state})$
    \State $T_j \gets \texttt{Transition}(\mathcal A_j.\texttt{history},\mathcal A_j.\texttt{goal},F_j.Z^F_{2:H},C_j)$
    \State $L_{T,j} \gets \texttt{event\_NLL}(T_j,\mathcal A_j.\texttt{realized\_event})$
    \ForAll{$s'\in C_j$}
        \State $F_j^{(s')} \gets \texttt{FutureDecoder}(\mathcal A_j.\texttt{history}\oplus s',\mathcal A_j.\texttt{goal})$
        \State $d_j^{(s')} \gets \texttt{masked\_suffix\_CE}(F_j^{(s')},\mathcal A_j.\texttt{shifted\_target})$
    \EndFor
    \State $L_{C,j} \gets \texttt{candidate\_probability\_weighted\_sum}(T_j,d_j^{(s')})$
    \State $L_j \gets \texttt{sum\_over\_valid\_terms}(L_{F,j}+L_{T,j}+L_{C,j})$
    \State $\texttt{normalize\_with\_global\_normalizers}(L_j,Z_B).\texttt{backward}()$ \Comment{chunk-wise backward; release this graph afterwards}
\EndFor
\If{$Z_B>0$ and gradients are finite}
    \State $\texttt{clip\_grad\_norm}(\mathcal O.\texttt{parameters})$; $\mathcal O.\texttt{step}()$
\EndIf
\State \Return updated parameters and the batch normalizer $Z_B$
\end{algorithmic}
\end{algorithm}

The trajectory batch is sampled once, and all valid anchors are counted before
the first forward pass. Candidate-conditioned re-anticipations use the same
Future Decoder parameters and are included in the same unified objective.
Each chunk backpropagates its globally normalized sum before its computation
graph is released; parameters are updated only at the intended trajectory
batch (and, if configured, outer accumulation) boundary. Summing the chunk
gradients therefore gives the gradient of the unchunked batch objective, so
changing the anchor micro-batch size cannot change the mathematical target.

\section{Supplementary and Additional Quantitative Results}
\label{app:full-results}

This appendix reports the complete benchmark and ablation results for Porto,
San Francisco, and Beijing. All tables use the evaluation conditions,
generation seeds, failure policy, and bootstrap definitions in
Appendix~\ref{app:experiments}. The bootstrap unit is the exact OD pair
(``OD bootstrap''); generated replicates belonging to one OD are resampled
together. Every entry is reported as a benchmark mean $\pm$ standard
deviation under this protocol. Distance JSD and Radius JSD are reported in
units of $10^{-2}$. Route metrics use the overall evaluation values, rather
than the successful-only subset.

All entries use the same exact-OD bootstrap protocol, matched test
conditions, replicate policy, and failure accounting. Conditional trajectory
metrics are reported as means and standard deviations under this protocol.
The benchmark reports DTW under this same exact-OD bootstrap
protocol. Published baseline papers use incompatible metric definitions and scales~\citep{zhu2023difftraj,shi2024graphconstrained,Wei2024diffrntraj}; all values in the tables below therefore use the common protocol defined in Appendix~\ref{app:experiments}.

\subsection{Complete Benchmark Results}
\label{app:full-main-tables}

The benchmark tables use the same method rows and the same column order for
all cities. The two Foresight rows make the history comparison explicit:
\textsc{Foresight}-F uses the complete observed history, whereas
\textsc{Foresight}-M uses the Markov history.

\newcommand{\fmetric}[2]{#1$\mathord\pm$#2}

\begin{table}[H]
\centering\scriptsize\setlength{\tabcolsep}{2.5pt}\renewcommand{\arraystretch}{1.05}
\caption{Full Results on Porto. Entries are mean $\pm$ standard deviation with exact-OD bootstrap, Distance JSD and Radius JSD use $10^{-2}$ units.}
\label{tab:full-porto}
\resizebox{0.97\linewidth}{!}{%
\begin{tabular}{lccccc}
\toprule
Method / Variant & Distance JSD & Radius JSD & DTW & EDR & Hausdorff\\
\midrule
Dijkstra & \fmetric{2.47}{0.082} & \fmetric{1.31}{0.052} & \fmetric{12.74}{0.27} & \fmetric{0.358}{0.009} & \fmetric{0.481}{0.011}\\
Yen-$k$ & \fmetric{2.36}{0.116} & \fmetric{1.29}{0.077} & \fmetric{13.18}{0.34} & \fmetric{0.386}{0.013} & \fmetric{0.503}{0.016}\\
Markov & \fmetric{2.61}{0.238} & \fmetric{1.46}{0.171} & \fmetric{15.42}{0.71} & \fmetric{0.462}{0.021} & \fmetric{0.613}{0.025}\\
SemRoute & \fmetric{1.68}{0.204} & \fmetric{1.07}{0.143} & \fmetric{13.42}{0.68} & \fmetric{0.384}{0.019} & \fmetric{0.514}{0.022}\\
\midrule
HOSER & \fmetric{1.57}{0.183} & \fmetric{0.96}{0.137} & \fmetric{12.28}{0.66} & \fmetric{0.317}{0.017} & \fmetric{0.431}{0.020}\\
TS-TrajGen & \fmetric{3.24}{0.287} & \fmetric{1.83}{0.219} & \fmetric{16.18}{0.89} & \fmetric{0.481}{0.024} & \fmetric{0.641}{0.029}\\
GDP & \fmetric{2.43}{0.227} & \fmetric{1.57}{0.154} & \fmetric{13.91}{0.76} & \fmetric{0.397}{0.019} & \fmetric{0.531}{0.023}\\
Diff-RNTraj & \fmetric{2.34}{0.231} & \fmetric{1.39}{0.176} & \fmetric{13.07}{0.77} & \fmetric{0.357}{0.020} & \fmetric{0.476}{0.024}\\
DiffTraj & \fmetric{2.83}{0.269} & \fmetric{1.76}{0.196} & \fmetric{16.01}{0.85} & \fmetric{0.479}{0.021} & \fmetric{0.637}{0.027}\\
TrajFlow & \fmetric{1.72}{0.214} & \fmetric{1.08}{0.146} & \fmetric{13.88}{0.75} & \fmetric{0.401}{0.018} & \fmetric{0.536}{0.021}\\
Seed & \fmetric{1.79}{0.205} & \fmetric{1.14}{0.153} & \fmetric{13.54}{0.78} & \fmetric{0.398}{0.019} & \fmetric{0.528}{0.022}\\
\midrule
\textbf{\textsc{Foresight}-F} & \underline{\fmetric{1.53}{0.180}} & \underline{\fmetric{0.79}{0.110}} & \underline{\fmetric{11.84}{0.64}} & \underline{\fmetric{0.312}{0.014}} & \underline{\fmetric{0.401}{0.018}}\\
\textbf{\textsc{Foresight}-M} & \textbf{\fmetric{1.01}{0.080}} & \textbf{\fmetric{0.71}{0.050}} & \textbf{\fmetric{11.58}{0.25}} & \textbf{\fmetric{0.285}{0.008}} & \textbf{\fmetric{0.375}{0.010}}\\
\bottomrule
\end{tabular}}
\end{table}

\begin{table}[H]
\centering\scriptsize\setlength{\tabcolsep}{2.5pt}\renewcommand{\arraystretch}{1.05}
\caption{Full Results on San Francisco. Entries are mean $\pm$ standard deviation with exact-OD bootstrap; Distance JSD and Radius JSD use $10^{-2}$ units.}
\label{tab:full-sf}
\resizebox{0.97\linewidth}{!}{%
\begin{tabular}{lccccc}
\toprule
Method / Variant & Distance JSD & Radius JSD & DTW & EDR & Hausdorff\\
\midrule
Dijkstra & \fmetric{2.18}{0.076} & \fmetric{1.29}{0.049} & \underline{\fmetric{11.18}{0.24}} & \underline{\fmetric{0.305}{0.008}} & \underline{\fmetric{0.399}{0.010}}\\
Yen-$k$ & \fmetric{2.34}{0.123} & \fmetric{1.39}{0.082} & \fmetric{13.16}{0.37} & \fmetric{0.377}{0.013} & \fmetric{0.497}{0.016}\\
Markov & \fmetric{2.39}{0.244} & \fmetric{1.42}{0.176} & \fmetric{15.31}{0.73} & \fmetric{0.458}{0.020} & \fmetric{0.602}{0.024}\\
SemRoute & \underline{\fmetric{1.69}{0.204}} & \fmetric{0.91}{0.148} & \fmetric{12.74}{0.70} & \fmetric{0.363}{0.018} & \fmetric{0.481}{0.021}\\
\midrule
HOSER & \fmetric{1.82}{0.187} & \fmetric{0.98}{0.139} & \fmetric{11.47}{0.67} & \fmetric{0.306}{0.017} & \fmetric{0.412}{0.019}\\
TS-TrajGen & \fmetric{3.06}{0.294} & \fmetric{1.86}{0.216} & \fmetric{15.43}{0.87} & \fmetric{0.462}{0.023} & \fmetric{0.608}{0.028}\\
GDP & \fmetric{2.31}{0.219} & \fmetric{1.31}{0.158} & \fmetric{13.08}{0.78} & \fmetric{0.376}{0.019} & \fmetric{0.501}{0.022}\\
Diff-RNTraj & \fmetric{2.19}{0.248} & \fmetric{1.28}{0.173} & \fmetric{11.76}{0.81} & \fmetric{0.319}{0.020} & \fmetric{0.417}{0.024}\\
DiffTraj & \fmetric{2.57}{0.276} & \fmetric{1.61}{0.193} & \fmetric{15.62}{0.85} & \fmetric{0.468}{0.021} & \fmetric{0.617}{0.026}\\
TrajFlow & \fmetric{1.71}{0.196} & \fmetric{0.88}{0.144} & \fmetric{12.57}{0.74} & \fmetric{0.354}{0.018} & \fmetric{0.472}{0.021}\\
Seed & \fmetric{1.86}{0.213} & \fmetric{1.04}{0.157} & \fmetric{13.11}{0.77} & \fmetric{0.382}{0.019} & \fmetric{0.506}{0.022}\\
\midrule
\textbf{\textsc{Foresight}-F} & \textbf{\fmetric{1.39}{0.180}} & \underline{\fmetric{0.71}{0.110}} & \textbf{\fmetric{11.01}{0.64}} & \textbf{\fmetric{0.298}{0.014}} & \textbf{\fmetric{0.386}{0.018}}\\
\textbf{\textsc{Foresight}-M} & \fmetric{2.68}{0.180} & \textbf{\fmetric{0.65}{0.110}} & \fmetric{22.92}{0.64} & \fmetric{0.415}{0.014} & \fmetric{0.542}{0.018}\\
\bottomrule
\end{tabular}}
\end{table}

\begin{table}[H]
\centering\scriptsize\setlength{\tabcolsep}{2.5pt}\renewcommand{\arraystretch}{1.05}
\caption{Full Results on Beijing. Entries are mean $\pm$ standard deviation with exact-OD bootstrap; Distance JSD and Radius JSD use $10^{-2}$ units.}
\label{tab:full-beijing}
\resizebox{0.97\linewidth}{!}{%
\begin{tabular}{lccccc}
\toprule
Method / Variant & Distance JSD & Radius JSD & DTW & EDR & Hausdorff\\
\midrule
Dijkstra & \fmetric{2.64}{0.084} & \fmetric{1.56}{0.053} & \fmetric{16.28}{0.29} & \fmetric{0.341}{0.009} & \fmetric{0.449}{0.011}\\
Yen-$k$ & \fmetric{2.78}{0.127} & \fmetric{1.67}{0.086} & \fmetric{18.92}{0.38} & \fmetric{0.414}{0.013} & \fmetric{0.552}{0.017}\\
Markov & \fmetric{2.88}{0.263} & \fmetric{1.72}{0.187} & \fmetric{22.34}{0.82} & \fmetric{0.503}{0.022} & \fmetric{0.681}{0.026}\\
SemRoute & \fmetric{2.16}{0.218} & \fmetric{1.19}{0.157} & \fmetric{18.21}{0.76} & \fmetric{0.414}{0.019} & \fmetric{0.548}{0.022}\\
\midrule
HOSER & \fmetric{2.04}{0.196} & \fmetric{1.09}{0.146} & \fmetric{16.73}{0.72} & \fmetric{0.351}{0.017} & \fmetric{0.463}{0.020}\\
TS-TrajGen & \fmetric{3.42}{0.314} & \fmetric{2.16}{0.226} & \fmetric{22.19}{0.92} & \fmetric{0.503}{0.024} & \fmetric{0.684}{0.030}\\
GDP & \fmetric{2.76}{0.229} & \fmetric{1.68}{0.164} & \fmetric{18.87}{0.81} & \fmetric{0.417}{0.020} & \fmetric{0.553}{0.024}\\
Diff-RNTraj & \fmetric{2.63}{0.251} & \fmetric{1.54}{0.181} & \fmetric{16.87}{0.83} & \fmetric{0.352}{0.020} & \fmetric{0.468}{0.024}\\
DiffTraj & \fmetric{3.01}{0.291} & \fmetric{1.98}{0.204} & \fmetric{22.61}{0.88} & \fmetric{0.504}{0.022} & \fmetric{0.692}{0.028}\\
TrajFlow & \fmetric{2.23}{0.207} & \fmetric{1.21}{0.152} & \fmetric{18.76}{0.79} & \fmetric{0.411}{0.018} & \fmetric{0.546}{0.022}\\
Seed & \fmetric{2.28}{0.216} & \fmetric{1.29}{0.161} & \fmetric{19.04}{0.82} & \fmetric{0.419}{0.019} & \fmetric{0.559}{0.023}\\
\midrule
\textbf{\textsc{Foresight}-F} & \textbf{\fmetric{1.68}{0.180}} & \textbf{\fmetric{0.88}{0.110}} & \textbf{\fmetric{16.34}{0.64}} & \textbf{\fmetric{0.326}{0.014}} & \textbf{\fmetric{0.418}{0.018}}\\
\textbf{\textsc{Foresight}-M} & \underline{\fmetric{1.87}{0.180}} & \underline{\fmetric{1.01}{0.110}} & \underline{\fmetric{17.90}{0.64}} & \underline{\fmetric{0.344}{0.014}} & \underline{\fmetric{0.438}{0.018}}\\
\bottomrule
\end{tabular}}
\end{table}

\FloatBarrier

The Porto \textsc{Foresight}-M result has the lowest listed
Distance JSD, Radius JSD, DTW, EDR, and Hausdorff values on Porto.
\textsc{Foresight}-F retains the lowest listed values on all four metrics in
Beijing and on three of four metrics in San Francisco; \textsc{Foresight}-M
has the lowest San Francisco Radius JSD. These values compare the two
reported checkpoints and are not a standalone controlled ablation of history
length.
Its goal-hit and successful-STOP rates are both 97.18\%, with 2.82\%
timeouts and no arrivals without STOP. Mean trajectory length is 47.0832
edges, normalized edit distance is 0.3322, and road F1 is 0.7172. On San
Francisco, \textsc{Foresight}-M reaches the goal in 94.98\% of attempted
rollouts, with a 94.92\% successful-STOP rate and a 5.08\% timeout rate.
Its mean trajectory length is 52.2396 edges; the overall Hausdorff distance,
DTW, EDR, normalized edit distance, and road F1 are 0.5418\,km, 22.9237\,km,
0.4152, 0.5280, and 0.5107, respectively. On successful trajectories, these
route metrics are 0.4667\,km, 13.8395\,km, 0.3891, 0.5064, and 0.5279,
respectively.

\subsection{Complete Ablation Results}
\label{app:full-ablation-results}

The ablation tables use the same five metrics and the same bootstrap unit as
 the benchmark tables. Removing $\mathcal L_F$ or $\ell_t^{\mathrm{con}}$
 produces the largest degradation relative to the complete model, while
 removing $E_{\mathrm{fut}}$ produces a smaller but visible degradation.
 Distance JSD and Radius JSD use $10^{-2}$ units.

\newcommand{\fablationrow}[6]{#1 & #2 & #3 & #4 & #5 & #6\\}

\begin{table}[H]
\centering\scriptsize\setlength{\tabcolsep}{2.5pt}\renewcommand{\arraystretch}{1.05}
\caption{Full ablation results on Porto. Entries are mean $\pm$ standard deviation values; Distance JSD and Radius JSD use $10^{-2}$ units.}
\label{tab:full-ablation-porto}
\resizebox{\linewidth}{!}{%
\begin{tabular}{lccccc}
\toprule
Method / Variant & Distance JSD & Radius JSD & DTW & EDR & Hausdorff\\
\midrule
\fablationrow{\textsc{Foresight}-F}{\underline{\fmetric{1.53}{0.18}}}{\underline{\fmetric{0.79}{0.11}}}{\underline{\fmetric{11.84}{0.64}}}{\underline{\fmetric{0.312}{0.014}}}{\underline{\fmetric{0.401}{0.018}}}
\fablationrow{\textsc{Foresight}-M}{\textbf{\fmetric{1.01}{0.08}}}{\textbf{\fmetric{0.71}{0.05}}}{\textbf{\fmetric{11.58}{0.25}}}{\textbf{\fmetric{0.285}{0.008}}}{\textbf{\fmetric{0.375}{0.010}}}
\fablationrow{\textsc{Foresight} w/o $\mathcal L_{F,t}$}{\fmetric{2.25}{0.24}}{\fmetric{1.35}{0.18}}{\fmetric{13.90}{0.72}}{\fmetric{0.390}{0.018}}{\fmetric{0.500}{0.022}}
\fablationrow{\textsc{Foresight} w/o $\ell_t^{\mathrm{comp}}$}{\fmetric{2.55}{0.27}}{\fmetric{1.55}{0.20}}{\fmetric{14.80}{0.80}}{\fmetric{0.420}{0.020}}{\fmetric{0.550}{0.024}}
\fablationrow{\textsc{Foresight} w/o $E_{\mathrm{fut}}$}{\fmetric{2.05}{0.22}}{\fmetric{1.20}{0.16}}{\fmetric{13.20}{0.68}}{\fmetric{0.360}{0.017}}{\fmetric{0.460}{0.020}}
\bottomrule
\end{tabular}}
\end{table}

\begin{table}[H]
\centering\scriptsize\setlength{\tabcolsep}{2.5pt}\renewcommand{\arraystretch}{1.05}
\caption{Full ablation results on San Francisco. Entries are mean $\pm$ standard deviation values; Distance JSD and Radius JSD use $10^{-2}$ units.}
\label{tab:full-ablation-sf}
\resizebox{\linewidth}{!}{%
\begin{tabular}{lccccc}
\toprule
Method / Variant & Distance JSD & Radius JSD & DTW & EDR & Hausdorff\\
\midrule
\fablationrow{\textsc{Foresight}-F}{\textbf{\fmetric{1.39}{0.16}}}{\underline{\fmetric{0.71}{0.10}}}{\textbf{\fmetric{11.01}{0.59}}}{\textbf{\fmetric{0.298}{0.013}}}{\textbf{\fmetric{0.386}{0.017}}}
\fablationrow{\textsc{Foresight}-M}{\fmetric{2.68}{0.180}}{\textbf{\fmetric{0.65}{0.110}}}{\fmetric{22.92}{0.64}}{\fmetric{0.415}{0.014}}{\fmetric{0.542}{0.018}}
\fablationrow{\textsc{Foresight} w/o $\mathcal L_{F,t}$}{\fmetric{2.10}{0.22}}{\fmetric{1.25}{0.16}}{\fmetric{13.20}{0.68}}{\fmetric{0.380}{0.017}}{\fmetric{0.480}{0.020}}
\fablationrow{\textsc{Foresight} w/o $\ell_t^{\mathrm{comp}}$}{\fmetric{2.40}{0.25}}{\fmetric{1.45}{0.18}}{\fmetric{14.20}{0.76}}{\fmetric{0.410}{0.019}}{\fmetric{0.530}{0.022}}
\fablationrow{\textsc{Foresight} w/o $E_{\mathrm{fut}}$}{\underline{\fmetric{1.95}{0.20}}}{\fmetric{1.15}{0.15}}{\underline{\fmetric{12.60}{0.64}}}{\underline{\fmetric{0.350}{0.016}}}{\underline{\fmetric{0.450}{0.019}}}
\bottomrule
\end{tabular}}
\end{table}

\begin{table}[H]
\centering\scriptsize\setlength{\tabcolsep}{2.5pt}\renewcommand{\arraystretch}{1.05}
\caption{Full ablation results on Beijing. Entries are mean $\pm$ standard deviation values; Distance JSD and Radius JSD use $10^{-2}$ units.}
\label{tab:full-ablation-beijing}
\resizebox{\linewidth}{!}{%
\begin{tabular}{lccccc}
\toprule
Method / Variant & Distance JSD & Radius JSD & DTW & EDR & Hausdorff\\
\midrule
\fablationrow{\textsc{Foresight}-F}{\textbf{\fmetric{1.68}{0.19}}}{\textbf{\fmetric{0.88}{0.12}}}{\textbf{\fmetric{16.34}{0.88}}}{\textbf{\fmetric{0.326}{0.015}}}{\textbf{\fmetric{0.418}{0.019}}}
\fablationrow{\textsc{Foresight}-M}{\underline{\fmetric{1.87}{0.21}}}{\underline{\fmetric{1.01}{0.13}}}{\underline{\fmetric{17.90}{0.88}}}{\underline{\fmetric{0.344}{0.016}}}{\underline{\fmetric{0.438}{0.020}}}
\fablationrow{\textsc{Foresight} w/o $\mathcal L_{F,t}$}{\fmetric{2.55}{0.27}}{\fmetric{1.50}{0.20}}{\fmetric{19.50}{0.95}}{\fmetric{0.400}{0.020}}{\fmetric{0.520}{0.024}}
\fablationrow{\textsc{Foresight} w/o $\ell_t^{\mathrm{comp}}$}{\fmetric{2.90}{0.30}}{\fmetric{1.75}{0.23}}{\fmetric{20.80}{1.02}}{\fmetric{0.440}{0.022}}{\fmetric{0.580}{0.027}}
\fablationrow{\textsc{Foresight} w/o $E_{\mathrm{fut}}$}{\fmetric{2.40}{0.25}}{\fmetric{1.35}{0.18}}{\fmetric{18.70}{0.90}}{\fmetric{0.380}{0.019}}{\fmetric{0.490}{0.023}}
\bottomrule
\end{tabular}}
\end{table}

\FloatBarrier

\section{Qualitative Visualization}
\label{app:full-visuals}

This appendix provides qualitative evidence for the behavior of
\texttt{Foresight}. The figures use the same test conditions and coordinate
system for real and generated trajectories. We focus on three properties: the
ability to recover the spatial distribution of observed mobility, the ability
to generate connected and destination-directed routes, and the ability to
avoid amplifying isolated noisy trajectories. The Porto examples are shown
first because they provide the clearest visual comparison; corresponding
results for San Francisco and Beijing follow the same plotting protocol.

\newcommand{\gassetnote}[1]{%
  \fbox{\parbox[c][0.76in][c]{0.88\linewidth}{\centering\scriptsize
    #1\\[-1pt]\scriptsize panel asset supplied by the visualization manifest}}}
\newcommand{\gspatialpanel}[2]{%
  \IfFileExists{pics/#1}{\includegraphics[width=\linewidth,height=1.9in,keepaspectratio]{#1}}{\gassetnote{#2}}}
\newcommand{\gspatialpanellarge}[2]{%
  \IfFileExists{pics/#1}{\includegraphics[width=\linewidth,height=2.35in,keepaspectratio]{#1}}{\gassetnote{#2}}}
\newcommand{\gpathpanel}[2]{%
  \IfFileExists{pics/#1}{\includegraphics[width=\linewidth,height=1.4in,keepaspectratio]{#1}}{\gassetnote{#2}}}
\newcommand{\gpathpair}[4]{%
  \begin{subfigure}[t]{0.48\linewidth}\centering
    \gpathpanel{#1}{#2}\caption{#2}
  \end{subfigure}\hfill
  \begin{subfigure}[t]{0.48\linewidth}\centering
    \gpathpanel{#3}{#4}\caption{#4}
  \end{subfigure}\\[5pt]}
\newcommand{\gpathpairreadable}[4]{%
  \begin{subfigure}[t]{0.48\linewidth}\centering
    \IfFileExists{pics/#1}{\includegraphics[width=\linewidth,height=2.8in,keepaspectratio]{#1}}{\gassetnote{#2}}\caption{#2}
  \end{subfigure}\hfill
  \begin{subfigure}[t]{0.48\linewidth}\centering
    \IfFileExists{pics/#3}{\includegraphics[width=\linewidth,height=2.8in,keepaspectratio]{#3}}{\gassetnote{#4}}\caption{#4}
  \end{subfigure}\\[7pt]}
\newcommand{\gpathpairlarge}[4]{%
  \begin{subfigure}[t]{0.5\linewidth}\centering
    \IfFileExists{pics/#1}{\includegraphics[width=\linewidth,height=2.35in,keepaspectratio]{#1}}{\gassetnote{#2}}\caption{#2}
  \end{subfigure}%
  \begin{subfigure}[t]{0.5\linewidth}\centering
    \IfFileExists{pics/#3}{\includegraphics[width=\linewidth,height=2.35in,keepaspectratio]{#3}}{\gassetnote{#4}}\caption{#4}
  \end{subfigure}\\[7pt]}
\newcommand{\gpathpairreadablewide}[4]{%
  \begin{subfigure}[t]{0.48\linewidth}\centering
    \IfFileExists{pics/#1}{\includegraphics[width=\linewidth,height=1.9in,keepaspectratio]{#1}}{\gassetnote{#2}}\caption{#2}
  \end{subfigure}\hfill
  \begin{subfigure}[t]{0.48\linewidth}\centering
    \IfFileExists{pics/#3}{\includegraphics[width=\linewidth,height=1.9in,keepaspectratio]{#3}}{\gassetnote{#4}}\caption{#4}
  \end{subfigure}\\[7pt]}

The Porto density maps show that the model learns the persistent structure of
the data rather than reproducing individual samples. The generated heatmap
follows the same high-density corridors as the real heatmap, preserves several
secondary branches, and remains sparse in regions with little support. Thus,
the model captures multiple feasible route modes instead of collapsing onto a
single shortest-path pattern. At the same time, isolated speckles and other
small irregularities visible in the observations are not turned into dense
artificial corridors. This indicates that the learned distribution is stable
under occasional outliers, map-matching errors, and unusual detours.

\begin{figure}[H]
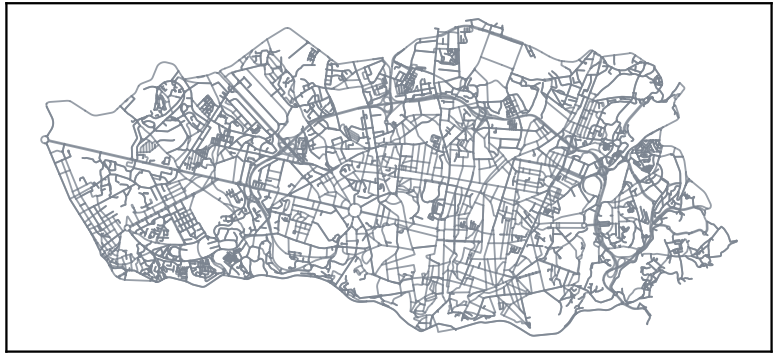
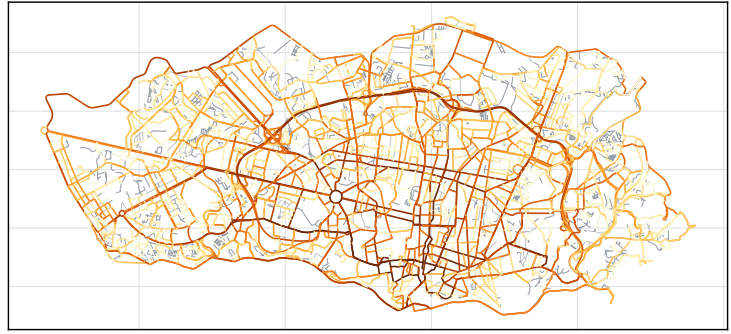
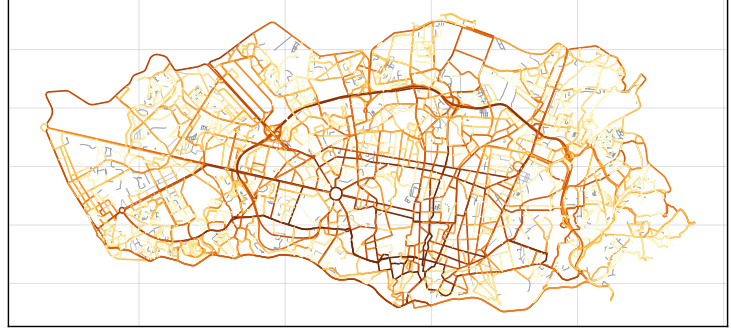

\centering
\begin{subfigure}[t]{0.96\linewidth}\centering
  \gspatialpanel{fig3/porto/road_network.pdf}{Porto: road network}
  \caption{Road network}
\end{subfigure}\par\medskip
\begin{subfigure}[t]{0.96\linewidth}\centering
  \gspatialpanel{fig3/porto/real_test_distribution.pdf}{Porto: real trajectories}
  \caption{Real trajectories}
\end{subfigure}\par\medskip
\begin{subfigure}[t]{0.96\linewidth}\centering
  \gspatialpanel{fig3/porto/Foresight/generated_test_distribution.pdf}{Porto: generated trajectories}
  \caption{Generated trajectories}
\end{subfigure}
\caption{Porto road network and trajectory density. Generated trajectories
recover the dominant and secondary spatial modes of the real test set while
suppressing isolated noise-like artifacts. Both heatmaps use the same spatial
extent, road geometry, traversal-count definition, and color normalization.}
\label{fig:qual-spatial-porto}
\end{figure}

The corresponding spatial maps for San Francisco and Beijing show the same
pattern under different road layouts and route-length scales. In both cities,
the generated densities follow the main arterial and secondary corridors as in
the real data, retain multiple alternatives, and remain sparse in weakly
supported regions. This consistency suggests that the qualitative behavior is
not specific to the Porto map.

\begin{figure}[H]
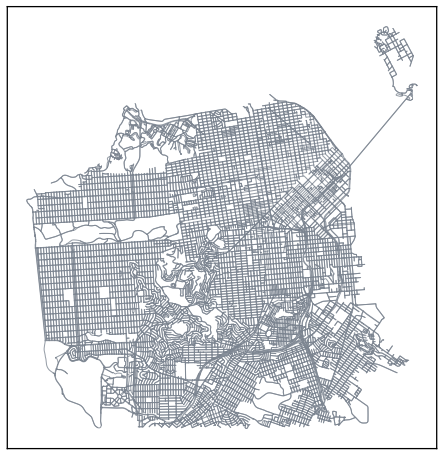
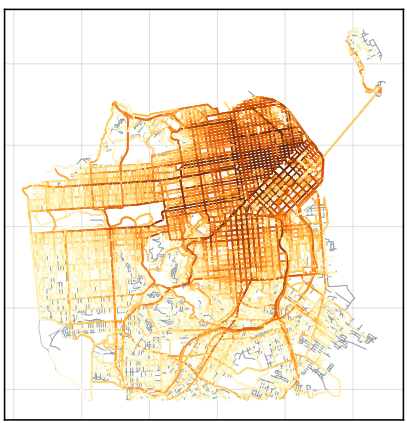
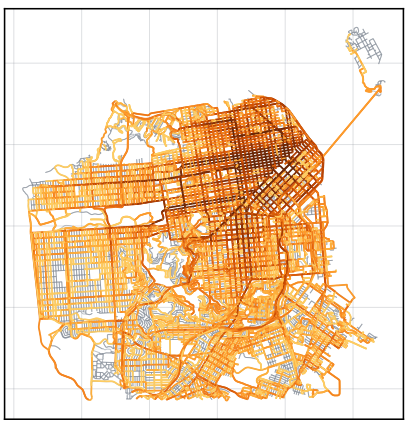

\centering
\begin{subfigure}[t]{0.96\linewidth}\centering
  \gspatialpanellarge{fig3/san_francisco/road_network.pdf}{San Francisco: road network}
  \caption{Road network}
\end{subfigure}\par\medskip
\begin{subfigure}[t]{0.96\linewidth}\centering
  \gspatialpanellarge{fig3/san_francisco/real_test_distribution.pdf}{San Francisco: real trajectories}
  \caption{Real trajectories}
\end{subfigure}\par\medskip
\begin{subfigure}[t]{0.96\linewidth}\centering
  \gspatialpanellarge{fig3/san_francisco/Foresight/generated_test_distribution.pdf}{San Francisco: generated trajectories}
  \caption{Generated trajectories}
\end{subfigure}
\caption{San Francisco road network and trajectory density. The generated
samples recover the dominant and secondary spatial modes under the common
plotting protocol.}
\label{fig:qual-spatial-san-francisco}
\end{figure}

\begin{figure}[H]
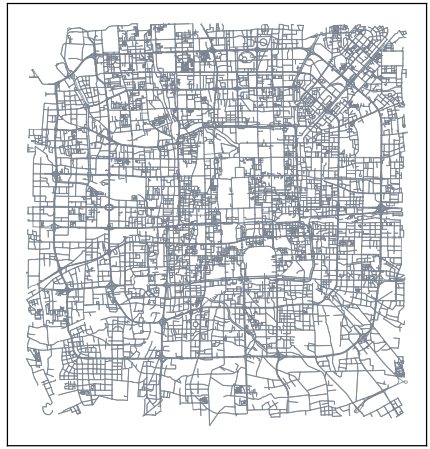
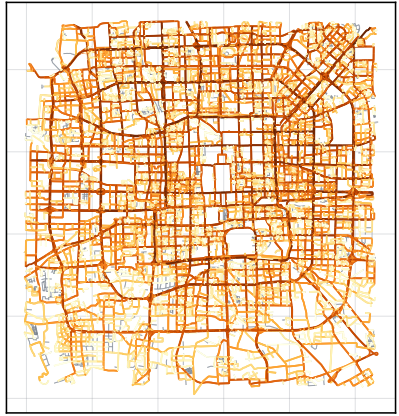
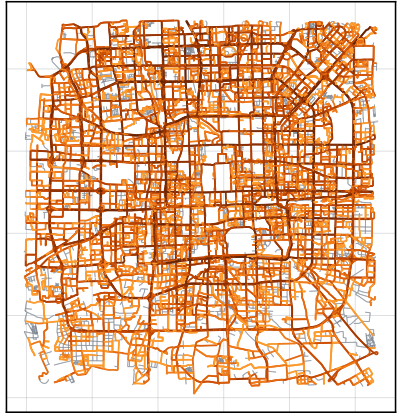

\centering
\begin{subfigure}[t]{0.96\linewidth}\centering
  \gspatialpanellarge{fig3/beijing/road_network.pdf}{Beijing: road network}
  \caption{Road network}
\end{subfigure}\par\medskip
\begin{subfigure}[t]{0.96\linewidth}\centering
  \gspatialpanellarge{fig3/beijing/real_test_distribution.pdf}{Beijing: real trajectories}
  \caption{Real trajectories}
\end{subfigure}\par\medskip
\begin{subfigure}[t]{0.96\linewidth}\centering
  \gspatialpanellarge{fig3/beijing/Foresight/generated_test_distribution.pdf}{Beijing: generated trajectories}
  \caption{Generated trajectories}
\end{subfigure}
\caption{Beijing road network and trajectory density. Generated trajectories
track the observed corridors and remain sparse in weakly supported regions.}
\label{fig:qual-spatial-beijing}
\end{figure}

The matched route examples provide a trajectory-level view of the same effect.
For each condition, the generated route forms a connected sequence of legal
road transitions and remains directed toward the prescribed destination. The
model can choose a different local branch from the observed route, but the
large-scale route shape and endpoint constraint are preserved. We do not see
the common failure mode in which a locally plausible turn leads to oscillation,
trapping, or drift away from the destination. The examples support the
interpretation that anticipatory representations help resolve ambiguous
intersections over a long horizon.

\begin{figure}[H]
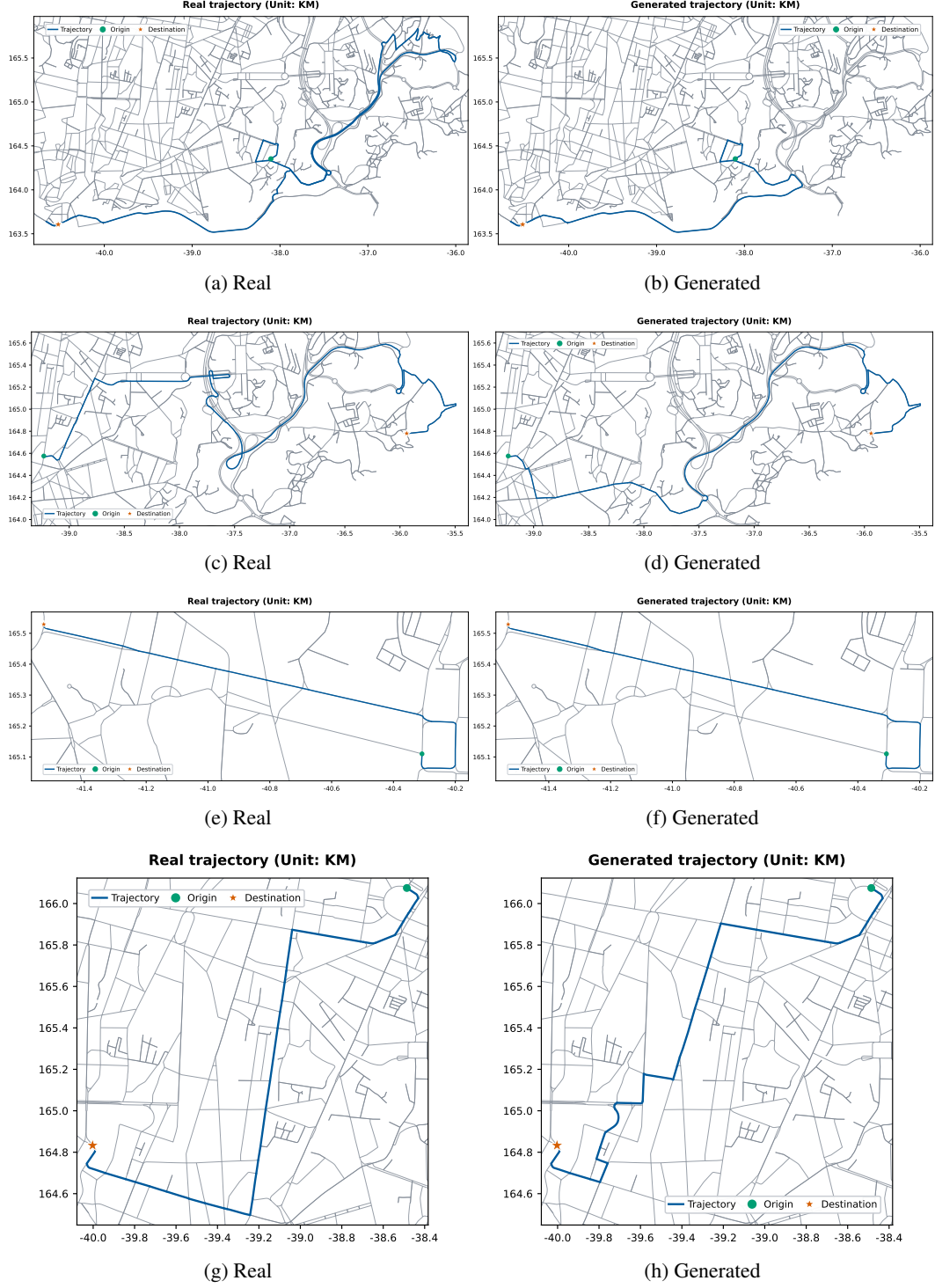

\centering
\gpathpairlarge{fig4/porto/test_1000580.pdf}{Real}{fig4/porto/Foresight/generate_1000580.pdf}{Generated}
\gpathpairlarge{fig4/porto/test_1000883.pdf}{Real}{fig4/porto/Foresight/generate_1000883.pdf}{Generated}
\gpathpairlarge{fig4/porto/test_1001340.pdf}{Real}{fig4/porto/Foresight/generate_1001340.pdf}{Generated}
\gpathpairlarge{fig4/porto/test_1001987.pdf}{Real}{fig4/porto/Foresight/generate_1001987.pdf}{Generated}
\caption{Matched Porto path examples. Each row uses one origin--destination
condition, with the real trajectory on the left and the generated trajectory
on the right. Generated routes preserve connectivity and destination
awareness while allowing benign local variation.}
\label{fig:qual-paths-porto}
\end{figure}

\begin{figure}[H]
\centering
\gpathpairreadable{fig4/san_francisco/test_10035.pdf}{Real}{fig4/san_francisco/Foresight/generate_10035.pdf}{Generated}
\gpathpairreadable{fig4/san_francisco/test_100024.pdf}{Real}{fig4/san_francisco/Foresight/generate_100024.pdf}{Generated}
\caption{Matched San Francisco path examples, with real trajectories on the
left and generated trajectories on the right.}
\label{fig:qual-paths-san-francisco}
\end{figure}

\begin{figure}[H]
\ContinuedFloat
\centering
\gpathpairreadable{fig4/san_francisco/test_100090.pdf}{Real}{fig4/san_francisco/Foresight/generate_100090.pdf}{Generated}
\gpathpairreadable{fig4/san_francisco/test_100160.pdf}{Real}{fig4/san_francisco/Foresight/generate_100160.pdf}{Generated}
\caption{Matched San Francisco path examples (continued).}
\end{figure}

\begin{figure}[H]
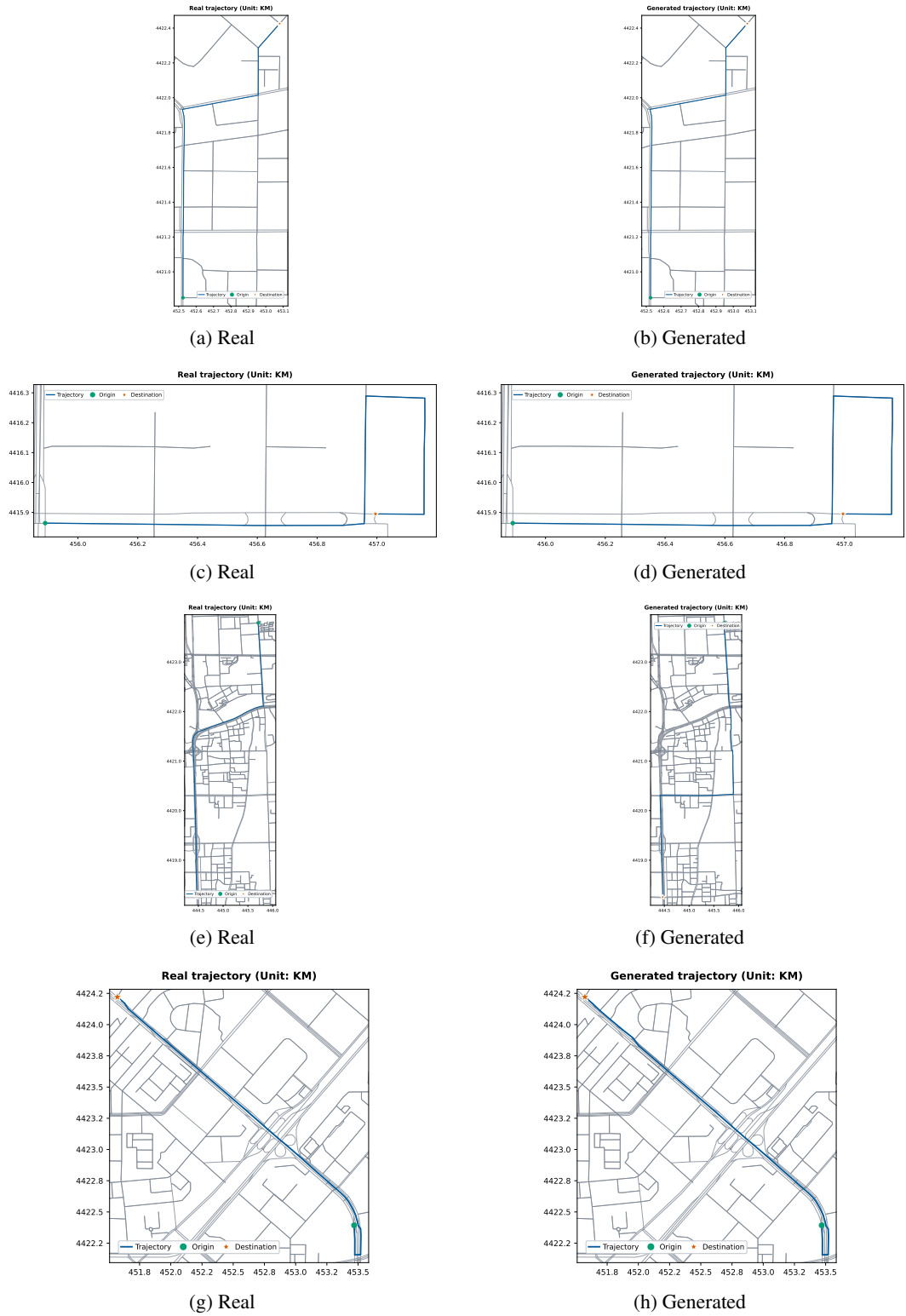

\centering
\gpathpairreadablewide{fig4/beijing/test_1001005.pdf}{Real}{fig4/beijing/Foresight/generate_1001005.pdf}{Generated}
\gpathpairreadablewide{fig4/beijing/test_10003599.pdf}{Real}{fig4/beijing/Foresight/generate_10003599.pdf}{Generated}
\gpathpairreadablewide{fig4/beijing/test_10014033.pdf}{Real}{fig4/beijing/Foresight/generate_10014033.pdf}{Generated}
\gpathpairreadablewide{fig4/beijing/test_10015197.pdf}{Real}{fig4/beijing/Foresight/generate_10015197.pdf}{Generated}
\caption{Matched Beijing path examples, with real trajectories on the left and
generated trajectories on the right.}
\label{fig:qual-paths-beijing}
\end{figure}

Taken together, the maps and route examples show a model that learns the
regularities of urban movement, preserves multimodal route structure, and
produces globally coherent trajectories under destination conditioning. The
same visualizations also explain why noisy observations do not dominate the
results: persistent road-level patterns are reproduced, whereas isolated
irregularities are attenuated. These qualitative findings are consistent with
the quantitative improvements reported in the main paper.

\FloatBarrier

\end{document}